\documentclass[lettersize,journal]{IEEEtran}
\usepackage{amsmath,amsfonts,amssymb}
\usepackage{algorithmic}
\usepackage{algorithm}
\usepackage{array}
\usepackage[caption=false,font=normalsize,labelfont=sf,textfont=sf]{subfig}
\usepackage{textcomp}
\usepackage{stfloats}
\usepackage{url}
\usepackage{verbatim}
\usepackage{graphicx}
\usepackage{cite}
\usepackage{multirow}

\usepackage{textcomp}
\usepackage{hyperref}
\usepackage{wrapfig}
\usepackage{booktabs}
\usepackage{float}

\usepackage[framemethod=tikz]{mdframed}

\begin{document}

\title{Advancing 6-DoF Instrument Pose Estimation in Variable X-Ray Imaging Geometries}

\author{Christiaan G.A. Viviers,~\IEEEmembership{Graduate Student Member, IEEE} Lena Filatova, Maurice Termeer, Peter H.N. de With,~\IEEEmembership{Life Fellow, IEEE} and Fons van der Sommen,~\IEEEmembership{Member, IEEE}


\thanks{C.G.A. Viviers is with the Technical University of Eindhoven, Eindhoven, the Netherlands (e-mail: c.g.a.viviers@tue.nl) and in collaboration with Philips IGT, 5684 PC, Best, the Netherlands (e-mail: partner.christiaan.viviers@philips.com).}
\thanks{Maurice Termeer and  Lena Filatova are with Philips IGT, 5684 PC, Best, the Netherlands (e-mail:maurice.termeer@philips.com; lena.filatova@philips.com).}
\thanks{P.H.N. de With and F. van der Sommen are with the Electrical Engineering Department,
Technical University of Eindhoven, Endhoven, The Netherlands (e-mail: P.H.N.de.With@tue.nl; fvdsommen@tue.nl).}}

\markboth{Author Version - IEEE TRANSACTIONS ON IMAGE PROCESSING, VOL. 33, 2024}%
{Viviers \MakeLowercase{\textit{et al.}}: Advancing 6-DoF Instruments Pose Estimation in Variable X-Ray Imaging Geometries}

\IEEEpubid{Author Version - 10.1109/TIP.2024.3378469~\copyright~2024 IEEE}

\maketitle

\begin{abstract}

Accurate 6-DoF pose estimation of surgical instruments during minimally invasive surgeries can substantially improve treatment strategies and eventual surgical outcome. Existing deep learning methods have achieved accurate results, but they require custom approaches for each object and laborious setup and training environments often stretching to extensive simulations, whilst lacking real-time computation. We propose a general-purpose approach of data acquisition for 6-DoF pose estimation tasks in X-ray systems, a novel and general purpose YOLOv5-6D pose architecture for accurate and fast object pose estimation and a complete method for surgical screw pose estimation under acquisition geometry consideration from a monocular cone-beam X-ray image. The proposed YOLOv5-6D pose model achieves competitive results on public benchmarks whilst being considerably faster at 42~FPS on GPU. In addition, the method generalizes across varying X-ray acquisition geometry and semantic image complexity to enable accurate pose estimation over different domains. Finally, the proposed approach is tested for bone-screw pose estimation for computer-aided guidance during spine surgeries. The model achieves a 92.41\% by the 0.1$\cdot d$ ADD-S metric, demonstrating a promising approach for enhancing surgical precision and patient outcomes. The code for YOLOv5-6D is publicly available at \href{https://github.com/cviviers/YOLOv5-6D-Pose}{https://github.com/cviviers/YOLOv5-6D-Pose}.
\end{abstract}


\section{Introduction}
\label{sec:introduction}

\IEEEPARstart{F}{luoroscopy-guided} minimally invasive interventions have greatly improved patient outcome from trauma, orthopedic or cancer surgeries. These image-guided surgeries largely rely on repeated acquisition of standard projections for instrument guidance and monitoring. Instrument maneuvering is typically performed manually by the clinician's hand (through trial and error) and without additional assistance, requiring multiple and extended sessions of fluoroscopy at the expense of additional radiation to the patient. Procedures are complex and due to an often very limited spatial configuration, surgical results are error-prone and highly surgeon-dependent. 

Recently, various methods have been proposed to improve instrument positioning during interventional surgeries. Expertise in interventional radiology and image guidance has expanded the treatment options for bone surgeries such as pedicle screws placement in the thoracic and lumbosacral spine~\cite{Learch2004-zg}. State-of-the-art (SOTA) practice for pedicle screw placement employs an intraoperative cone-beam computed tomography~(CBCT) scan and combines it with an external navigation system. The intra-operative 3D augmented reality surgical navigation (ARSN) system uses external optical video cameras to augment the surgical field and assist the clinician in the navigation path for screw placement. Screw placement is then confirmed with an additional postoperative CT scan~\cite{aug_screw_1,aug_screw_2} and manual validation. In line with previous approaches in this field~\cite{brainlab}, this comes at the expense of extensive external equipment and alters the clinical way of working, which inhibits adoption. Although these methods demonstrate progress in screw placement by indicating a path, they do not provide any guidance or validation through actual screw tracking. 

\IEEEpubidadjcol
Providing surgical guidance by extracting semantic information from the X-ray images alone is extremely appealing with benefits for several applications. Cardiac interventions have utilized this and improved the visualization of both catheter-based devices and soft tissue anatomy by co-registering X-ray fluoroscopy~(XRF) images with echocardiography through Transesophageal echocardiography~(TEE) probe pose estimation from the X-ray image alone~\cite{cardiac_tee}. Screw placement surgery is another example of a complex procedure that can greatly benefit from extracting visual information available in the X-ray image for surgical assistance. Through pose estimation via accurate 6 Degrees of Freedom~(DoF) of the surgical instruments from a single X-ray image, additional guidance to clinicians is provided during image-guided procedures and instrument placement is determined without the need for additional external navigation systems or postoperative CT scans.


Motivated by the need for automated robot operation, autonomous driving and VR \& AR applications, methods for accurate 6-DoF pose estimation of rigid objects have extensively been studied~\cite{review}. While most existing methods assume a fixed image acquisition geometry, which is sufficient for many applications, some domains, such as X-ray imaging or space satellite pose estimation, require the imaging geometry to constantly change during its operation. Adjustment of the focal length (zooming) or the detector field of view (X-ray image size and dose control) are common changes in such a framework. Naturally, it is also evident that pose estimation methods should include the intrinsic camera parameters if the methods will be used across different cameras or risk manual adjustment for each camera (re-training and data collection in case of learning-based methods). Therefore, in these domains it is crucial for pose estimation methods to incorporate the changing imaging geometry to accurately recover the pose of the target object(s). Perspective-n-Point (PnP) deep learning-based methods, which use the intrinsic camera parameters to estimate the object's 6-DoF pose, could be readily applied to these domains. However, the accuracy of these approaches, as originally proposed by Tekin~~\textit{et al.}~\cite{yolo6D}, was limited by the YOLOv2 architecture's inability to accurately regress 2D image locations of the projected vertices of the object's 3D bounding box. Other two-staged methods such as EPro-PnP~\cite{epro}, employ an initial object detection method followed the final object pose estimation, making them computationally less efficient. Recent advancements in the YOLO object detection series suggest that 6-DoF pose estimation can benefit from these improvements to efficiently achieve high pose accuracy under variable acquisition geometry.\\

In this work, we propose a general-purpose deep learning-based instrument pose-estimation method, addressing technical challenges that have limited such technology from being incorporated in practice. We build on our prior work~\cite{xray-6d-pose} and introduce a novel YOLOv5-6D pose architecture for more accurate and fast object 6-DoF pose estimation. In addition, to address the difficulty in acquiring data, a data collection method is introduced for automatic data labeling that generalizes across all cone-beam X-ray geometries and object types.

Our contributions are as follows.
\begin{itemize}
  \item A general-purpose approach for data acquisition for instrument 6-DoF pose estimation in X-ray imaging tasks is proposed. By utilizing an external optical camera, we can acquire the transformation between the instrument of interest and the X-ray system, an Azurion C-arm system (Philips IGT, Best, Netherlands). This allows for automated data acquisition and labeling, which is then possible by projecting the object onto the detector at any object or C-arm position. 
  \item A novel YOLOv5-based architecture for 6-DoF pose estimation is introduced for accurate and fast object pose estimation, hereafter referred to as YOLOv5-6D. This newly proposed method extends on recent advancements in the YOLO object detection series for the pose estimation task. The proposed approach achieves competitive results on the public RGB-based benchmarks (LINEMOD) and is faster than previous methods enabling real-time operation.
  \item To enable accurate 6-DoF pose estimation in X-ray, we include the X-ray imaging geometry in the estimation task. The proposed data acquisition method combined with YOLOv5-6D, achieves a strong baseline performance on the newly acquired asymmetrical calibration cube dataset of 99.27\% 0.1$\cdot d$ average distance difference (ADD).
  \item The method generalizes across imaging system geometries and to more complex imaging environments. We demonstrate this effectiveness and accuracy of 6-DoF pose estimation on a clinically-relevant cannulated cancellous screw and show that the YOLOv5-6D model is capable of generalizing outside of its training domain to a more complex setting, indicating its potential for real-world applications.
\end{itemize}

As a solution direction, the object pose is acquired through predicting 2D key points for the instrument's virtual 3D bounding-box and resolving the pose through a Perspective-n-Point~(PnP) algorithm~\cite{PnP} under consideration of the acquisition geometry. This attribute enables the transition to the X-ray domain, where the acquisition geometry is constantly changing during a procedure and across systems. Additionally, we address generalization from our training domain to a clinically relevant setting through a series of extensive augmentations. The proposed method shows robustness and high accuracy for 6-DoF pose estimation of a surgical screw in a variable intra-operative setting.\\ 

This paper is organized as follows. Section~\ref{sec:relatedwork} discusses the related work on object 6-DoF pose estimation for both the color and X-ray domain. Section~\ref{sec:approach} introduces the proposed approach to X-ray-based object pose estimation. The results of these experiments are presented in Section~\ref{sec:results}. Finally, a discussion on the obtained results and possible future directions are included in Section~\ref{sec:discussion} and ~\ref{sec:limitations}.

\section{Related Work}
\label{sec:relatedwork}
Recent advancements in deep learning have improved the accuracy at which systems can estimate the position (3-DoF) and orientation (3-DoF) of rigid objects. This progress is largely driven by applications for the metaverse, VR \& AR, robot operation and intelligent driving. Zhu~\textit{et al.}~\cite{review} provide an extensive review of methods for 6-DoF pose estimation. Extending this review, we briefly consider related work in object pose estimation in the RGB and X-ray domain. We omit a detailed discussion of methods dependent on depth information (RGB-D), such as RCVPose~\cite{wu2022vote} and PVN3D~\cite{he2020pvn3d}, as well as RGB-D-based, model-free methods like FS6D~\cite{Fs6d} and the more recent FoundationPose~\cite{wen2023foundationpose}, since this depth modality is unavailable in our X-ray setting.

\subsection{6-DoF Pose Estimation in RGB}
The majority of research efforts in object 6-DoF pose estimation determine the object pose from RGB images with knowledge of the object of interest. These methods commonly utilize the object 3D models followed by task-specific model training. More recently there has also been growing interest in generalizeble 3D object model free methods (hereafter refereed to as model-free methods), that do not require additional training to predict the pose of novel objects~\cite{gen6d, lin2023vi, wang2023query6dof}. While this does alleviate large constraints on employing the method, they do still fall behind in terms of both speed and accuracy (Table~\ref{fig:speed_tradeoff}). In many applications, with the object 3D model often obtainable, accuracy and speed is required over ease of implementation.

The state-of-the-art methods employing object specific knowledge during training can roughly be categorized as methods that (1)~directly regress object pose from the color image (referred to as \textit{direct} methods), (2)~employ a PnP algorithm to compute the object pose from 2D predicted key points of a corresponding 3D model (referred to as \textit{PnP} methods) and (3)~either (1)~or (2)~followed by an iterative refinement procedure.\\

Direct pose estimation involves directly regressing the object pose with, typically, a deep convolutional neural network~(CNN) from the RGB image in an end-to-end fashion. Bukschat~\textit{et al.} proposed EfficientPose~\cite{bukschat2020efficientpose} that employs the EfficientNet~\cite{efficientNet} backbone and a BiFPN-net~\cite{tan2020efficientdet} to regress the object pose from RGB images at different scales. EfficientPose regresses the pose of single objects from RGB images in the LINEMOD benchmark at 36.43 ms/image (27.45 FPS) and an average 97.35\% $0.1\cdot d$ ADD(-S) accuracy. Methods in this category do not explicitly consider the camera acquisition geometry and these parameters are thus considered static and fixed per camera/model pair. Recently, Xu~\textit{et al.} developed RNNPose~\cite{rnnpose} that starts with an initial pose from any method (tested with a direct~\cite{posecnn} and PnP method~\cite{pvnet}) and iteratively refines the object pose, based on the estimated correspondence field between the reference (2D render of 3D model) and target images. This iterative re-projection strategy considers the intrinsic camera parameters, but comes at the cost of increased computation time. This method currently achieves the highest accuracy on the public LINEMOD benchmark at 97.37\% ADD(-S), but with an inference time (4 rendering cycles and 4 recurrent iterations each as per the paper) of 308.35 ms/image (3.24 FPS), excluding the initial pose prediction step.

Tekin~\textit{et al.} proposed a 2D-3D correspondence-based method (SingleShotPose~\cite{yolo6D} also known as YOLO-6D) for 6-DoF pose estimation. The model simultaneously performs a single-shot object detection and the 6D pose prediction from an RGB image. This is realized by predicting the 2D image locations of the projected vertices of the object’s 3D bounding box. Using a Perspective-n-Point (PnP) algorithm and known acquisition parameters, the 6D pose of an object can be estimated (we mention the relationship here, but will discuss it in detail in Section~\ref{sec:approach}). As a feature extraction network, the model used the Darknet19-448 backbone, first proposed in YOLOv2~\cite{YOLOv2} for object detection. Since its release, there have been considerable improvements in YOLO object detection series~\cite{yolov3,yolov4,Wang_2021_CVPR,yolov5}. We leverage these advances as we develop the YOLOv5-6D pose estimation model. 

Prior to our work, other PnP-based 6-DoF pose estimation methods have been developed~\cite{yolo6D,pvnet}, with the best-performing method being EPro-PnP~\cite{epro}, proposed by Chen~\textit{et al.}. This two-staged approach achieves a high 96.36\% ADD(-S) accuracy, enabled by the dense correspondences extracted form the object image crop and the proposed differentiable PnP layer. The pose estimation step is also computationally efficient (see Section~\ref{sec:inference_time}). While the pose estimation step and the PnP layer can be integrated with any architecture, their work employs CDPN~\cite{li2019cdpn}, a dense correspondence network for 6-DoF pose estimation from object-specific image crops. An initial method is thus required for object detection and cropping on the target images which typically consumes majority of the compute budget. Section~\ref{sec:inference_time} provides an in-depth run-time analysis of these methods.

\subsection{Object Pose Estimation in X-ray}

Methods for 6-DoF pose estimation in the X-ray domain have been proposed for applications ranging from industrial product inspection, C-arm repositioning for surgical assistance, to surgical tool pose estimation. Presenti~\textit{et al.} propose a series of methods~\cite{Presenti2020, Presenti_CNN, PRESENTI2023118866} to recover manufactured object pose from X-ray images for defect inspection. Their approach assumes fixed acquisition geometry and displays sub-optimal results when only one image is used~\cite{PRESENTI2023118866}, compared to methods employing PnP. Similarly, X-Ray-PoseNet~\cite{x-ray-posnet} has been proposed by Bui~\textit{et al.} to directly regress the the translation (3~degrees) and rotation (4~quaternions) of industrial objects with respect to the X-ray system. Their approach is based on a custom CNN architecture and assumes fixed X-ray acquisition geometry, while being trained on simulated X-ray images. Kausch~\textit{et al.}~\cite{C-arm-positioning} developed a C-arm re-positioning pipeline to suggest C-arm imaging angles for assistance during spinal implant placement. It uses the patient spine as reference and suggests a new C-arm position through a series of features extracted from the X-ray image, using multiple U-Net-like models. Despite the interest in surgical tool guidance, few attempts have been made to directly recover the pose of the instrument used during the treatment. Registration between X-ray fluoroscopy (XRF) and transesophageal echocardiography~(TEE) for structural heart interventions relies on accurate pose-estimation of the TEE probe. TEE-probe pose estimation through 2D/3D registration methods based on iterative refinement such as Direct Splat Correlation~(DSC) and Patch Gradient Correlation~(PCG) have been implemented\cite{cardiac_tee}. Instrument pose estimation from 3D ultrasound data volumes has received substantially more attention~\cite{arashUltrasound, HongxuOverview}.

In one particular case, K\"{u}gler~\textit{et al.} developed i3PosNet~\cite{i3PosNet}, a method for surgical instruments pose estimation using a VGG~\cite{vgg}-based CNN architecture. The network predicts object-specific key points from localized patches. While considering the geometric landmarks of fiducials during pose estimation, the method does not account for the image acquisition geometry, limiting its application across different systems and geometries. i3PosNet is designed for pose estimation of symmetrical objects and lacks effectiveness for asymmetrical instruments as it only estimates a 5-DoF pose. This method is developed and trained on simulated data and finally tested on manually annotated real X-ray images which introduces time-consuming setup and potential human errors. Finally, the multistage approach employed, including image variety reduction, image information extraction followed by pose reconstruction from pseudo-landmarks, hinders its real-time applicability.

\subsection{Approaches for 6-DoF pose estimation Data Acquisition}
The LINEMOD dataset~\cite{linemod} is the most commonly used dataset for 6-DoF object pose estimation in the RGB(-D) domain. Images of the 15~objects are collected in sequence under different illumination and large viewpoint changes in a heavily cluttered environment with mild occlusions. The ground-truth poses (labels) are obtained using calibrated cameras and a calibration pattern. 

Unfortunately, labeled intra-operative X-ray training data for object 6-DoF pose estimation has neither been described nor published. All of the above-mentioned methods rely on simulated training data that require highly accurate simulations and extensive CAD modeling. The methods then train on these simulations, aiming to generalize to the test domain. This domain gap introduces a challenge when transferring to real-world applications. K\"{u}gler~\textit{et al.}~\cite{i3PosNet} acquire real data of object poses through tedious manual annotation effort, which involves projecting their object as an outline on the X-ray image and then interactively translating and rotating the object to match the X-ray image. The applied data in their approach is also captured using fixed X-ray acquisition geometry.\\

Summarizing the outcome of this detailed review, we position our work as a general-purpose 2D/3D correspondence method for instrument pose estimation from a single X-ray image. Building on our prior work~\cite{xray-6d-pose}, the method takes the X-ray acquisition geometry into account, enabling it to generalize to new systems. This generalization will be discussed in Section~\ref{sec:X-ray Pose Estimation}. To address the difficulty in data acquisition, we present a general method for capturing real X-ray data of any object.
\section{Approach}
\label{sec:approach}
\begin{figure*}[ht]
\centering
\includegraphics[width=14cm]{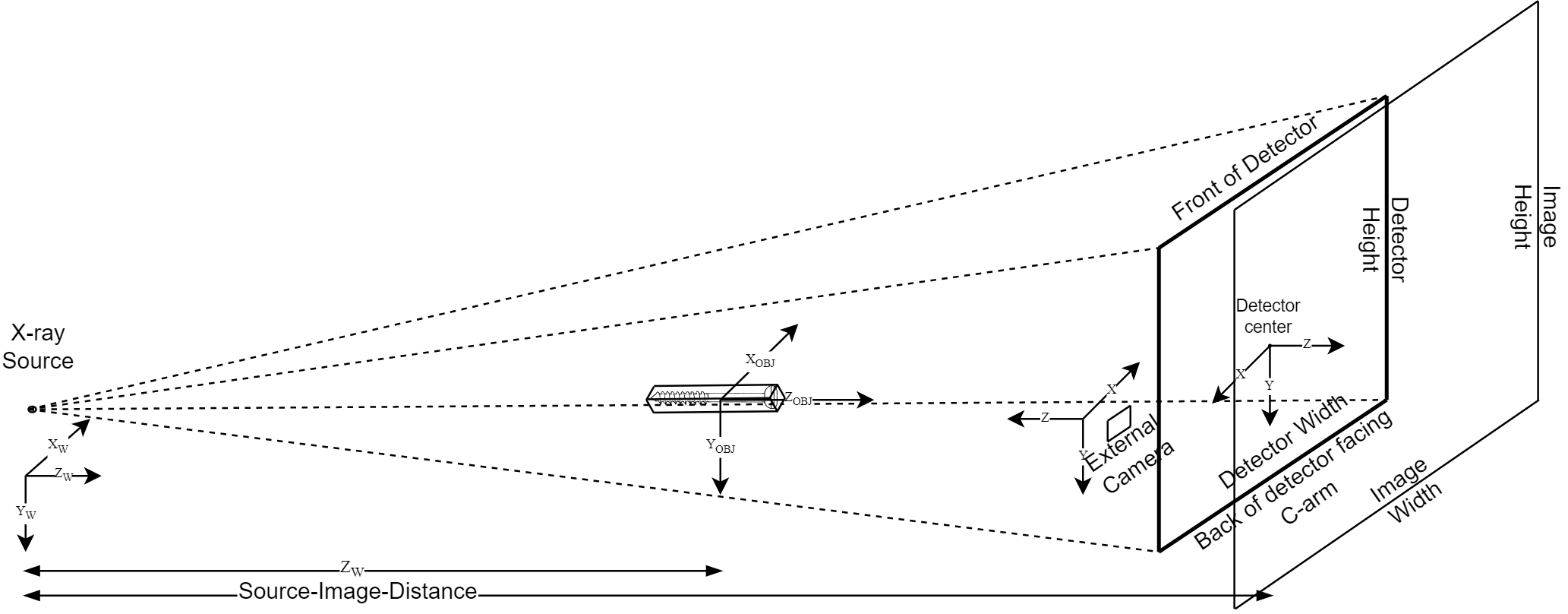}
\caption{X-ray projection model depicting the X-ray source, a surgical screw, detector with an attached grayscale optical camera, the detector panel and the captured X-ray image. The frame of reference for each point of interest is also depicted.}
\label{fig:xray_model}
\end{figure*} 
\subsection{X-ray Pose Estimation}\label{sec:X-ray Pose Estimation}
X-ray imaging systems are available in a range of different sizes with varying detector shapes, depending on the needs of the application. In addition, modern X-ray systems allow for the acquisition geometry to change at run-time to improve image quality of the area of interest. In brief, this results in varying acquisition parameters such as the detector size, detector field of view~(FOV) and most commonly, the source-image distance~(SID). All of these variables have a direct effect on the resulting X-ray image. Computer-aided image-guided methods influenced by these changes need to request fixed acquisition parameters or incorporate their variation in order to present accurate results. While several methods requesting fixed acquisition geometry have been adopted, they have limited applicability or require extensive additional preparation effort for each new system. Object pose estimation is fundamentally connected to the image acquisition parameters and, as such, we incorporate them in the proposed method to allow a single trained model to generalize to a wide range of acquisition geometries.  

\subsection{Data Acquisition Setup}\label{data acquisition}
Acquiring labeled data for 6-DoF pose estimation tasks is difficult, due to the inherent limitation of human observers to accurately determine an object's 6-DoF pose. When possible, manual labeling, even in the case of projected key points, is prone to errors and extremely laborious. Therefore, we draw inspiration from data collection methods in the optical domain~\cite{t-less, linemod} and devise a setup for accurate and automatic data acquisition and labeling for 6-DoF pose estimation in X-ray without corrupting (or introducing a learnable bias to) the X-ray image with external markers. 

In our data acquisition setup, we attach an external optical camera to the X-ray detector. The optical cues from the camera that are transparent to the X-ray, can be utilized to assist in the pose estimation task. The complete method consists of (1)~a ChArUco board~\cite{aruco} with our (2)~object of interest at a known location on the board, (3)~the optical camera capturing images of the board whilst (4)~the X-ray system captures X-ray images of the object. The 2D projection of the object's 3D bounding box onto the X-ray detector can then be acquired through the optical pose estimation of the ChArUco board and a series of frame transformations to the X-ray source coordinate system. This allows for fully automated data acquisition through automated movement of the X-ray C-arm and patient table to a diverse set of positions. In contrast to previous work, the method does not rely on accurate rotation or translation sensors from the X-ray system and can thus be used across a wider range of X-ray systems and still recover accurate data labels. In addition, the labeled images are void of any external cues that can be utilized to determine the object pose.

In this work, we employ OpenCV~\cite{opencv_library} for acquiring the pose of the board in the optical camera frame. This is achieved through the detection of the ChArUco markers in the optical image and combined with the knowledge of their physical flat-panel location on the printed ChArUco board on the patient table. Provided with the set of 2D-3D correspondences, the camera pose in the table coordinate system can be obtained by solving the PnP problem. 
\begin{figure}
\centering
\subfloat[Grayscale image of the ChArUco board and test cube on the patient table taken from the optical camera on the detector. The 2D projection of 3D cube outline can be seen in blue.]{\label{fig:charuco} {\includegraphics[width=1\linewidth, trim={0.0cm 0.0cm 0.0cm 0.0cm},clip]{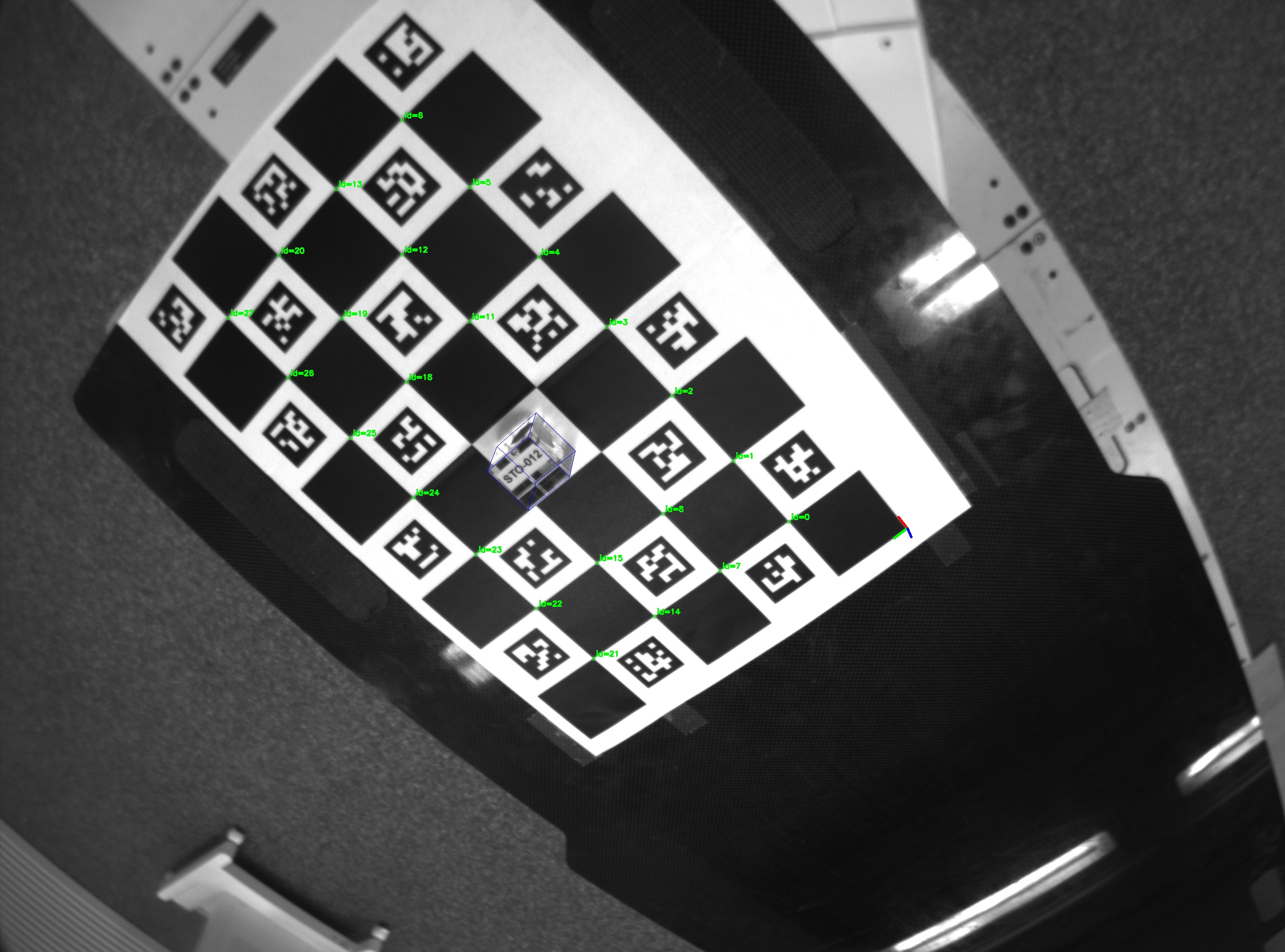}} }

\subfloat[Dicom cube image]{\label{fig:cube_dicom}{\includegraphics[width=0.5\linewidth]{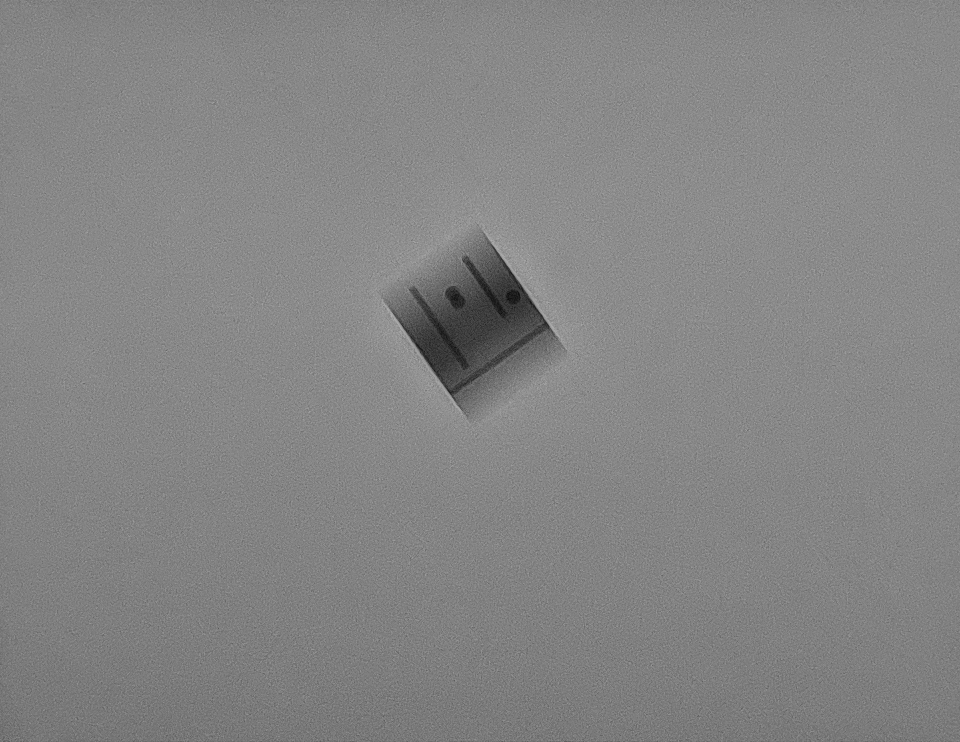}}}
\subfloat[Projected 3D bounding box]{\label{fig:cube_projection}{\includegraphics[width=0.5\linewidth, trim={0.0cm 0.1cm 0.0cm 0.0cm},clip]{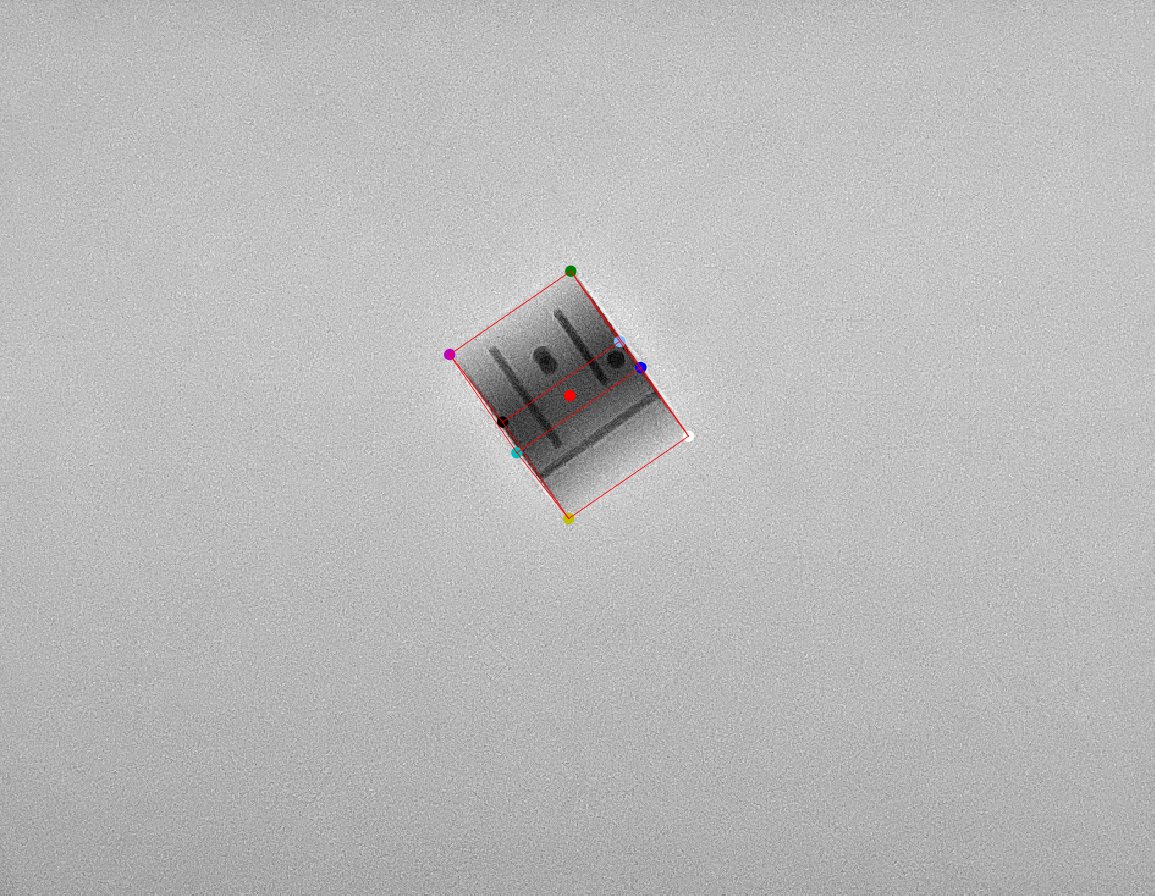}}}
\caption[Dataset acquisition setup]{(a) Grayscale image showcasing the setup used to automatically acquire the 6-DoF pose of various objects. (b) Corresponding X-ray Dicom image of the cube. (c) Projected 3D bounding box and virtual corner coordinates.}
\end{figure}

\subsection{X-ray Acquisition Model \& Calibration}\label{X-ray model}
In this work, we adopt the pinhole X-ray acquisition model to traverse between the 3D object frame and the 2D X-ray projection. The model is used during data acquisition and for the PnP pose calculation. Figure~\ref{fig:xray_model} depicts the X-ray acquisition model and Equation~(\ref{eq:projection_model}) formalizes it. The pinhole camera model is formally specified by

\begin{equation} \label{eq:projection_model}
\centering
\lambda \begin{bmatrix} u \\ v \\ 1 \end{bmatrix} = \begin{bmatrix} \textbf{K}  & \mathbf{O}_3 \end{bmatrix}
    \begin{bmatrix}
    \mathbf{R} & \mathbf{C} \\ \mathbf{O}_3^T & 1 
    \end{bmatrix}
    \begin{bmatrix}
    X_W \\ Y_W \\ Z_W \\ 1 
    \end{bmatrix},
\end{equation}
where the intrinsic parameter matrix $K$ is defined as 
\begin{equation}
\mathbf{K}=
\begin{bmatrix}
k_{u}f     & 0    & k_{u}x_{0} \\
0   & -k_{v}f  & k_{v}y_{0} \\
0   & 0    & 1 
\end{bmatrix}.
\label{eqn:K_params}
\end{equation}
In the above expressions, matrix $\mathbf{K}$ represents the intrinsic system parameters, which can change during the clinical operation and across different systems. These intrinsic parameters consist of the horizontal ($k_{u}$) and vertical ($k_{v}$) density of pixels. The pixel density can change depending on the detector and image size combination, e.g. changing the field of view or zooming on an image. The offset of the principal point to the detector center is represented by the coordinates ($x_{0}$, $y_{0}$). The source-image distance (SID), or focal length, is represented by parameter $f$. The extrinsic parameters $\mathbf{R}$ and $\mathbf{C}$ are the rotation and translation matrices to be solved. 
\begin{figure}[b]
\centering
\subfloat[Zoomed X-ray image of the surgical screw used for training and validation.]{\label{fig:plain_screw}{\includegraphics[width=0.485\linewidth, trim={2.0cm 2.0cm 3.2cm 2.0cm},clip]{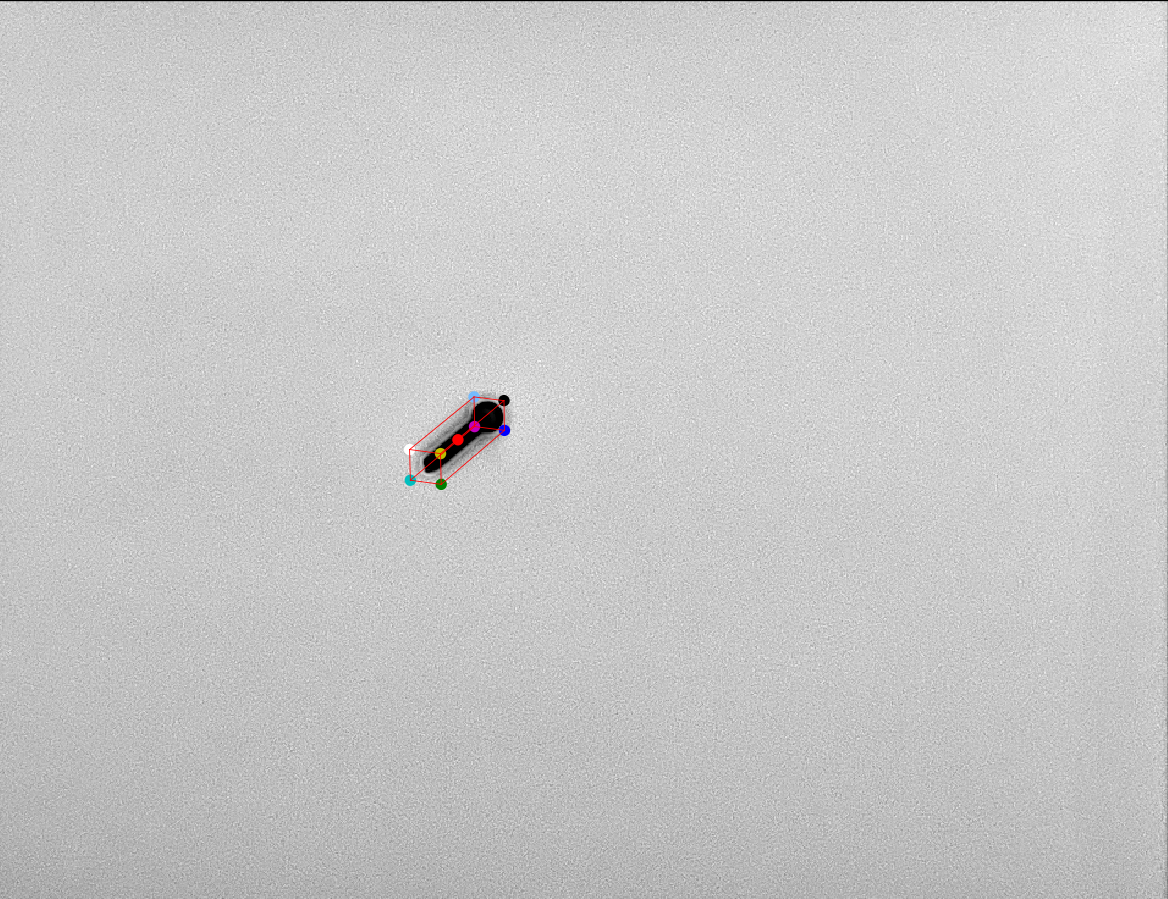}}}
\subfloat[X-ray image of the surgical screw with spine phantom used as test set.]{\label{fig:phantom_screw}{ \includegraphics[width=0.485\linewidth]{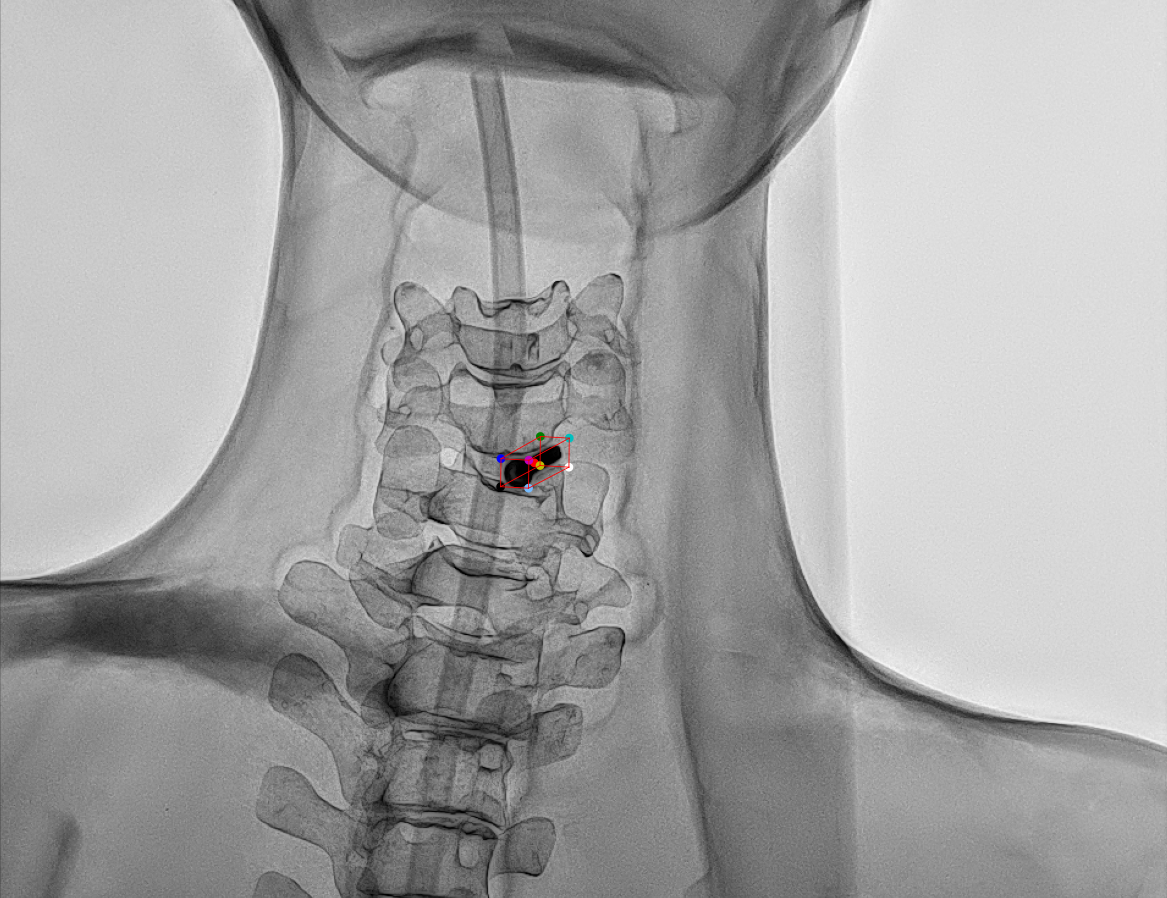}}}
\caption[Screw dataset]{Examples from the applied screw train, validation and test datasets. Each image also showcases the projected 3D bounding box of the screw.}
\end{figure}
The intrinsic parameters of the optical cameras are calibrated using a flat plate with a pattern of circles. Optical images of the plate are captured from different angles and the camera parameters are adjusted by minimizing the reprojection error of the circles. The resulting optical coordinate system is subsequently linked by capturing both optical and X-ray images of a dome-shaped calibration object, consisting of white plastic cylinders embedded in black foam. On the optical images, only the circular sides of the cylinders are visible, of which the center is computed. The X-ray images are used to create a 3D reconstruction, on which the cylinders are segmented and the same points as observed in the optical images are computed. The two point clouds are matched, thereby linking the optical and X-ray coordinate systems.

\subsection{Datasets}
\subsubsection{LINEMOD}
In line with previous work on 6-DoF pose estimation, we also evaluate the proposed YOLOv5-6D architecture on the popular LINEMOD dataset. The dataset consists of 13~different objects, each with approximately 1,200~images that are placed in various scenes. In this benchmark there is a predefined division scheme for training and test images which we also adopt. The training set varies between 15-30\% of the object's dataset, making it a very small fraction of the dataset. This aspect makes it particularly interesting and relevant for the medical domain, in the sense that approaches for this benchmark need to learn and generalize from scarce data.
 
\subsubsection{Cube}
Using our data collection method described in Section~\ref{data acquisition}, we have composed a dataset, henceforth referred to as the cube dataset, to test the adaptation of the proposed approach and the YOLOv5-6D network to the X-ray domain. Figure~\ref{fig:charuco} depicts the grayscale image of the $30\times30\times30$-mm perspex cube, embedded with metal markers placed on the ChArUco board. Since the 3D bounding box exactly matches that of the cube's physical dimensions, the cube is the ideal test object because one can visually determine the accuracy of the bounding box fit, whereas other objects might have a virtual 3D bounding box. Figures~\ref{fig:cube_dicom} and \ref{fig:cube_projection} depict the X-ray image of the cube and the corresponding 2D projection of the 3D bounding box. Along with the DICOM X-ray image and the 2D projected coordinates, we also capture the original 6-DoF cube pose and a binary mask of the cube in the X-ray image for training purposes. Table~\ref{capture params} lists the X-ray system's acquisition parameters and geometry used to capture the cube dataset, which is repeated for every side of the cube. In line with prior work~\cite{ADDS}, X-ray/optical image pairs are taken in 10-degree intervals across the geometrical rotation range of the X-ray system, to ensure a uniform viewing distribution of the object. The SID, translation and FOV parameters are uniformly sampled from the allowed range and automatic gain control manages the applied X-ray dose at a constant K Rate of 1.88 mGy/min. In total, we have acquired 1000 images ($r \in [-45^{\circ}, -35^{\circ}, -25^{\circ}, ..., +45^{\circ}] ^3 $) per cube side at a 960$\times$742 image resolution.

\begin{table}
\centering
\caption{C-arm acquisition and table parameters (w.r.t. its starting position) used during the data collection.}
\setlength{\tabcolsep}{3pt}
\begin{tabular}{l|l}
\toprule
\rule[-1ex]{0pt}{3.5ex}  Rotation  & Translation (mm)\\
\hline
\rule[-1ex]{0pt}{3.5ex}  $r_{z} \in  [-45^{\circ},45^{\circ}]$ & $t_{z} = 700 \pm 40 $ \\ 
\rule[-1ex]{0pt}{3.5ex} $r_{y} \in  [-45^{\circ},45^{\circ}]$ &  $ t_{y} = 0 \pm 40 $  \\ 
\rule[-1ex]{0pt}{3.5ex} $r_{x} \in [-45^{\circ},45^{\circ}]$ &  $t_{x} = 0 \pm 40 $   \\
\toprule
\rule[-1ex]{0pt}{3.5ex}  SID  (mm)  & FOV (mm) diagonal\\
\hline
\rule[-1ex]{0pt}{3.5ex}  $ [950.0, 1230.0]$   & $[156 , 484]$ \\	
\bottomrule
\end{tabular}
\label{capture params}
\end{table}

\subsubsection{Screws}
To demonstrate its clinical potential, we also evaluate the proposed approach for 6-DoF pose estimation of surgical screws for potential spine surgeries. The screw is a standard 3.5-mm cannulated cancellous screw often used during orthopedic surgeries. The screw is 34.3~mm long and has a head with a diameter of 6.88~mm. We have created a 3D model of the screw to be used during the projection onto the grayscale and X-ray image. The screw is inserted into a polystyrene block to enable precise placement on the ChArUco board. The same data collection method as described in Section~\ref{data acquisition} was followed to construct the screw dataset (Figure~\ref{fig:plain_screw}) for training and validation of the work. In addition to this dataset, we have also constructed a screw test dataset. The screw test dataset is set up to test the generalization of the proposed approach to a more realistic setting. We have attached the surgical screws to the spine of a human phantom, similar to their usage during a spine surgery. An example image of the screw and spine phantom can be seen in Figure~\ref{fig:phantom_screw}. While this setting is still different from an actual clinical intervention, it does enable us to determine whether the pose estimation method can generalize to a more complex domain. The screw and screw with human phantom dataset each contains 1000 images acquired following the parameters listed in Table~\ref{capture params} and as further specified in Section~\ref{clinical_context}.

\subsection{YOLOv5-6D Pose}
\begin{figure*}[t]
\centering
\includegraphics[width=\linewidth]{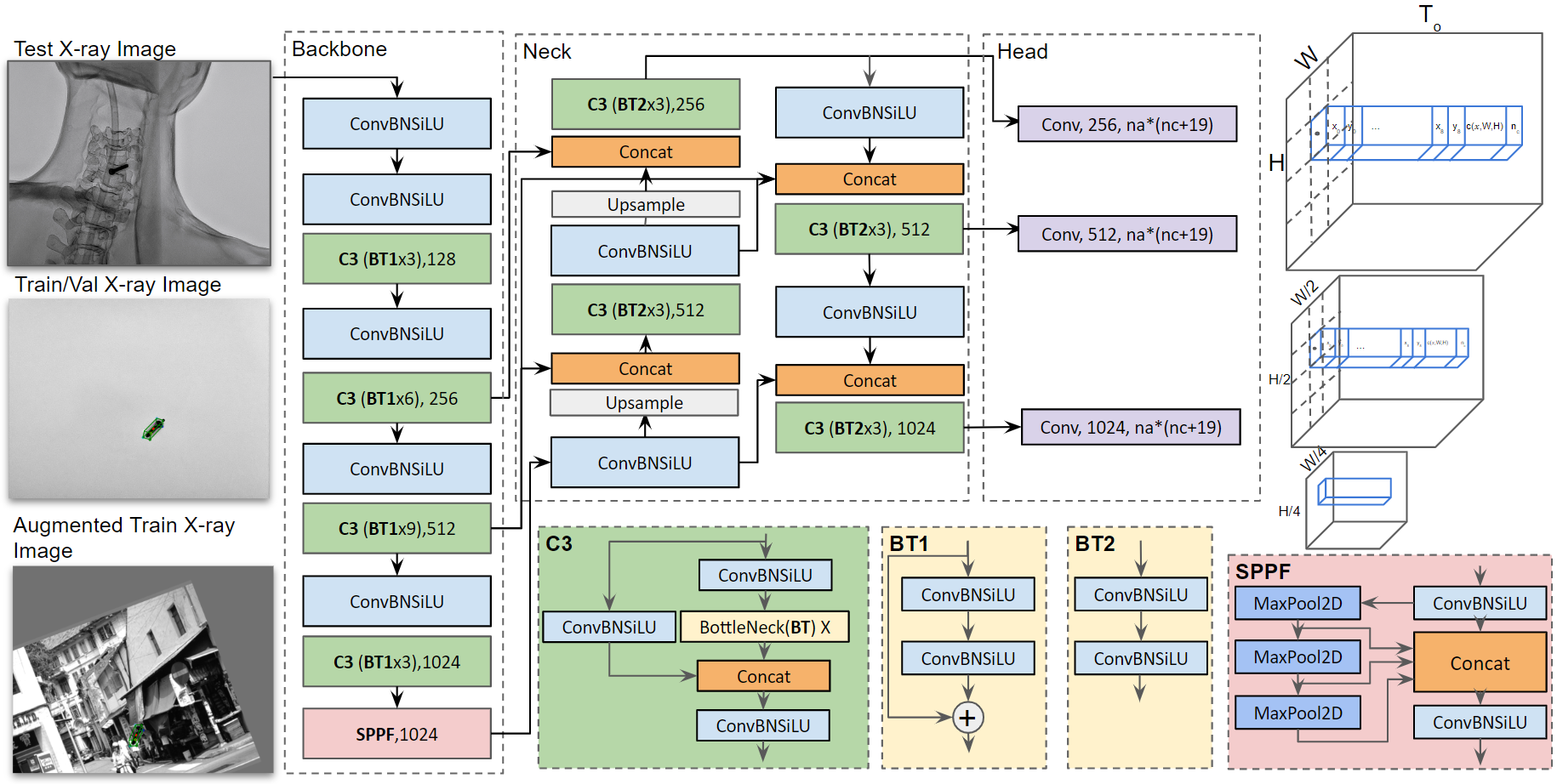}
\caption{Overview of the YOLOv5-6D architecture. The model backbone is based on the CSP-Net, the neck consists of the BiFPN architecture and the new model head predicts object key points at different scales. The model takes as input an image containing the object of interest and predicts object-specific key points that are used to estimate the object pose. The subblocks C3, BT1, BT2 and SPPF are depicted at the bottom in an enlarged view for further detail.}
\label{fig:architecture}
\end{figure*}

This research largely draws inspiration from the YOLO6D model~\cite{yolo6D} for object pose estimation and enhances it by incorporating recent advancements in the YOLO object detection series~\cite{yolov5}. As such, this single-shot approach enables simultaneous detection and 6-DoF pose estimation of objects in RGB and X-ray images. The model predicts the 2D image locations of the projected vertices of the object’s 3D bounding box. Using these 2D/3D correspondences and the current acquisition parameters, the object’s 6D pose is then solved using a PnP algorithm~\cite{PnP}, specifically ePnP~\cite{epnp} in our case. Figure~\ref{fig:architecture} depicts the YOLOv5-6D model architecture. The model follows a simple backbone, neck and head architecture. The backbone is based on the CSP-Net~\cite{CSP}, first proposed in the work by Wang~\textit{et al.} for improved object detection. For the model neck, the BiFPN~\cite{tan2020efficientdet} introduces a top-down pathway to fuse multi-scale features with an additional bottom-up pathway. The complete architecture shown in Figure~\ref{fig:architecture} can be sub-divided into different building blocks at a level of processing stages, indicated by different colors. These stages consist of (1)~ConvBNSilU - convolution, batch normalization, Silu activation, (2)~BottleNeck 1 - Two ConvBNSilU operations followed by a residual connection to the input, (3)~BottleNeck 2 - Two ConvBNSilU operations, (4)~C3 - ConvBNSilUs and a BottleNeck block (BT1 or BT2) (5)~SPFF - represents a pyramid structure through max pooling operations, (6) Conv - convolution.

We adjust the model head for key point prediction at different scales (three in our experiments). More precisely, three scales produce an $W\times H$ grid cells and $n_a$ anchor boxes (also three in our experiments) responsible for detecting the objects. Given the LINEMOD input images of size 640$\times$480, the network produces 18,900 predictions (80 × 60 × 3 + 40 × 30 × 3 + 20 × 15 × 3). Every cell and anchor-box combination predicts $\mathbf{T_o}$, which is the 2D location of the object center and 8~corners of the projected 3D bounding boxes in the image. More formally,  
$\mathbf{T_o} = (b_{x0}, b_{y0}), 8\times (b_x,  b_y), conf, n_{class} $, where $(b_{x0}, b_{y0})$ are the object center coordinates, $(b_x, b_y)$, the projected 3D bounding-box coordinates, $conf$ the cell confidence of it containing the object and $n_{class}$, the class-specific confidence. Hence, the model output comprises 19 predicted values, as we only capture one class. Additionally, we apply a scaled sigmoid function specified by
\begin{equation}\label{eq:center_coord}
f(\cdot) = (2(\sigma(\cdot)) - 0.5) + c_{\text{offset}}),
\end{equation}
to the object-center coordinates prediction for easier predictions when the object center is close to the edge of a grid cell compared to the original single sigmoid function. Finally, the prediction with the highest cell-specific object confidence is chosen for evaluation. In Equation~\ref{eq:center_coord}, $\sigma$ is the sigmoid activation function and $ c_{\text{offset}}$ is the offset to the top-left corner of the particular grid cell. 

In contrast to YOLO6D, our model incorporates a more advanced feature extraction backbone, utilizing CSP-Net over Darknet 19-448, and integrates an additional 'neck' network, BiFPN. This enhancement enables the feature extraction across multiple scales, as opposed to YOLO6D's single-scale approach. The features from these different scales are rasterized into the 18,900 cell predictions, compared to 845 cells in YOLO6D. This enables the network to make accurate predictions for much smaller and larger objects. These architectural improvements along with further refinements of the training objective leads to a significant accuracy increase at the cost of a minor speed decrease, as shown in Table~\ref{fig:speed_tradeoff}.

\subsection{Training Objective at Different Scales}
We introduce various technical improvements to enable efficient model training with the new architecture. The confidence function proposed by Tekin~\textit{et al.}~\cite{yolo6D} is adjusted to support the multi-scale model and variable input image dimensions. Most notably, we change the distance threshold used in the confidence function in Equation~(\ref{eq:confidence_function}), based on the output layer grid size instead of a fixed 2D Euclidean distance. The confidence function, $c(x, W, H)$, dynamically determines a grid cell's object confidence value for the current predicted 2D points based on its distance $D_T(x)$ from the target 2D points. Since the grid-space size changes at different output layers determined by the image resolution and aspect ratio, the confidence function is adjusted accordingly. This function can be formalized by
\begin{equation}\label{eq:confidence_function}
c(x, W, H) = \begin{cases}
      e^{\alpha(1-\frac{D(x)}{d_{T}(W, H)})}, & \text{if}\ D(x) < d_{T}(W, H) \\
      0, & \text{otherwise},
    \end{cases}
\end{equation}
where $ d_{T}(W, H) = \beta \sqrt{W^2 +H^2}$, the diagonal of the grid and $\beta$ a hyperparamter set to an empirically determined value of 0.2. The sharpness of the exponential function is determined by the hyperparameter $\alpha$ and $D(x)$ is the $L_1$ distance between the predicted point and the ground-truth point in grid-space coordinates. The complete loss function consists of $ \mathcal{L} = \lambda_{\text{points}}L_{\text{points}} + \lambda_{\text{conf}}L_{\text{conf}}$, where $\lambda_{\text{points}}$ and $ \lambda_{\text{conf}}$ are scaling hyperparameters to control the influence of the loss between points and the confidence loss, respectively.\newline

 The process of target prediction in our model involves a critical step of matching each target with the most suitable anchor, ensuring a close match between the widths of the target (determined by object-specific key points that are furthest apart - after augmentation - in the vertical and horizontal direction) and the anchor to determine the optimal scale for prediction. Following this, the model identifies the specific grid cell responsible for the prediction, based on the target's location. The primary cell for prediction is the one containing the target's center point, but adjacent cells may also participate, depending on the target's position within the cell. During training, grid cells are trained to predict targets in various positions. This process equips each grid cell with the ability to make accurate predictions for a range of target positions, thereby ensuring robustness and versatility in detecting different types of targets across various locations.

\subsection{Data Augmentation \& Training Details}\label{sec:training_details}
\begin{figure}[H]
\subfloat[X-ray cube image before any augmentation is applied.]{\label{fig:original_cube}{\includegraphics[width=0.5\linewidth]{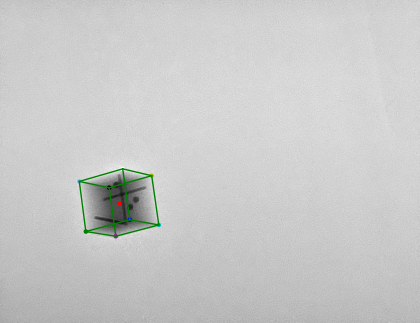}}}
\subfloat[The same image after augmentation.]{\label{fig:augmented_cube}{ \includegraphics[width=0.5\linewidth]{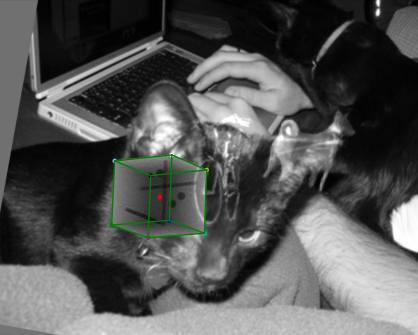}}}
\caption[augmentation]{Example image from the applied cube datasets from before and after augmentation for training. Notably, a bicycle partially ``occludes" the cube in the center of the augmented image, the background is changed and the image is scaled.}\label{fig:augmentation}
\end{figure}

We collect a small dataset of the object of interest and through extensive data augmentation, we establish the 6-DoF pose estimation YOLOv5-6D model to generalize to a new, unseen and more complex domain. A series of data augmentation techniques refined for accurate key point detection in both the RGB and X-ray domain are used. For the X-ray data loading pipeline, all data is processed as a one-channel image, compared to the normal three-channel RGB domain. The proposed augmentation pipeline consists of: (1) replacing the image background with a random image from the PASCAL VOC dataset~\cite{voc} using the object mask, (2) color-space HSV augmentation and contrast, brightness and noise adjustment for the grayscale images, (3) scaling (30\%), zooming ($\pm$ 30\%), translation (30\%), rotation ($\pm 180^{\circ}$) and sheering augmentation ($2^{\circ}$) and finally, (4) we employ an image overlay and occlusion strategy to randomly occlude (RGB images), or reduce the intensity (X-ray domain) of an area about or on top of the object of interest (X-ray ``occlusion"). Many of the occlusion augmentations were adapted from the work by S{\'a}r{\'a}ndi~\textit{et al.}~\cite{sarandi2018synthetic}. Figure~\ref{fig:augmentation} depicts this augmentation applied to an image from the cube dataset. The cube and screw datasets are randomly split into 70\%/30\% train/validation splits. 
YOLOv5-6D is branched from the YOLOv5 repository~\cite{yolov5} and adjusted for 6-DoF pose estimation instead of object detection. We utilize many of the training techniques in line with those used in object detection. We use an ADAM optimizer with a warm-up and cosine learning-rate scheduler. An $L_1$ loss is employed for key points and a cross-entropy loss for the objectiveness confidence. Model weights are initialized with the COCO-pretrained weights~\cite{cocodataset, yolov5} where possible. The above implementation is in PyTorch~1.7.0 and will be shared for reproducibility~\footnote{Code publicly available at: \url{https://github.com/cviviers/YOLOv5-6D-Pose}}. The models are trained on two RTX~3090Ti GPUs and all of the performance tests are carried out on a system with a more readily available RTX~2080Ti and an i9-9900KF CPU @ 3.60GHz for comparison.

\subsection{Evaluation Criteria}
We adopt the evaluation metrics from prior work on 6-DoF pose estimation. We employ the commonly used 3D distance of model vertices, often referred to as the average distance difference (ADD) and ADD-S (symmetric objects) metric~\cite{ADDS,brachmann2016uncertainty,kehl2017ssd}, as the main method of evaluation, while also providing further insight into model performance through the 2D reprojection error, average angle error and the translation error. The ADD metric can be equated as
\begin{equation}
\mathbf{ADD}=
\frac{1}{|\mathcal{M}|} \sum_{x\in \mathcal{M}} \lVert (\mathbf{R}x + \mathbf{t}) - (\bar{\mathbf{R}}x + \bar{\mathbf{t}}) \rVert_{2},
\label{eqn:ADD}
\end{equation}
and computes the average 3D distances between a set $\mathcal{M}$ of 3D points (the 3D model vertices) brought about the ground-truth rotation ($\mathbf{R}$) and translation ($\mathbf{t}$) and the predicted rotation ($\bar{\mathbf{R}}$) and translation ($\bar{\mathbf{t}}$). Averaging is done over the cardinality of $\mathcal{M}$. For symmetrical objects, we use the ADD-S metric defined as 
\begin{equation}
\mathbf{ADD\textbf{-} S}=
\frac{1}{|\mathcal{M}|} \sum_{x_1 \in \mathcal{M}} \underset{x_2 \in \mathcal{M}}{\min}\lVert (\mathbf{R}x_1 + \mathbf{t}) - (\bar{\mathbf{R}}x_2 + \bar{\mathbf{t}}) \rVert_{2},
\label{eqn:ADDS}
\end{equation}
capturing the smallest distance of the possible 3D distances. The 3D distance is converted into a binary metric based on a maximum object diameter threshold of 10\%, 5\% and 2\%. In Section~\ref{sec:inference_time} we also extensively measure and report model inference time.

\subsection{Clinical Context}\label{clinical_context}
During X-ray acquisition, attenuations along the beam direction are summed up and depth information is lost, potentially yielding ambiguous overlays of structures depending on the viewing direction. This is especially important when attempting to recover the pose of an instrument of interest. 
\begin{wrapfigure}{r}{0.20\textwidth}
\includegraphics[width=0.95\linewidth]{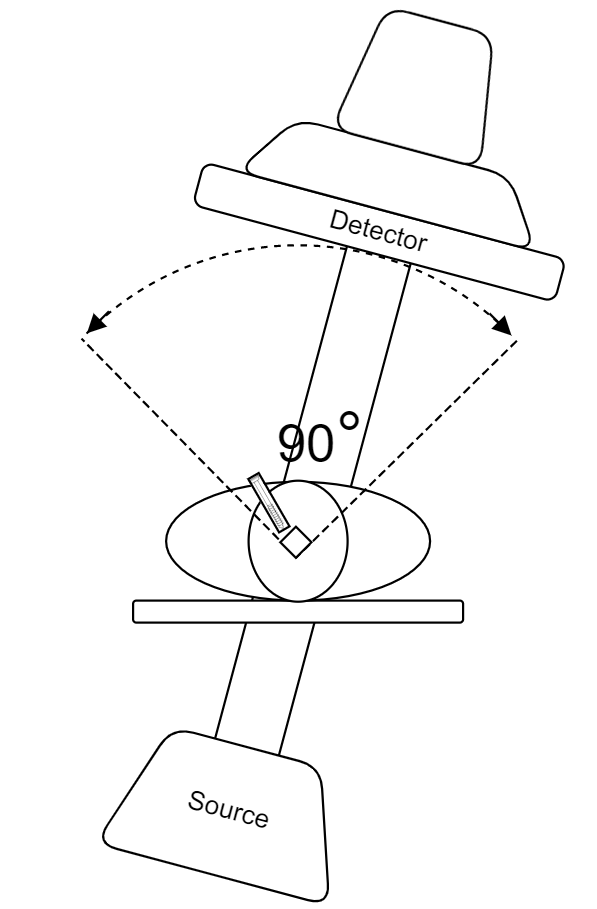} 
\caption{Typical C-arm working positions during spinal screw placement surgeries~\cite{aug_screw_1}.}
\label{fig:patient}
\end{wrapfigure}
Consequently, one can consider the deployment conditions and working positions of the X-ray system to determine if viewing angles can be constricted to avoid ambiguities, or if the ambiguous images even have an impact on the object pose (as in the case of symmetrical objects). With our application of spinal screw-placement surgeries in mind, the patient is typically in a prone positioning with the clinician performing the spinal surgery with a superior approach (from above the patient). Any screws being placed will be attached with the screw head upwards. This natural working condition can be exploited and ambiguities are directly avoided by limiting the viewing angles of the screw to be from above the patient as in Figure~\ref{fig:patient}. Although the rotational symmetry around the screw z-axis still remains, the physician will always check initial mounting to the correct vertebrae and will be concerned only about 5 degrees of freedom, explicitly the translation in $x,y,z$ (connecting point to the bone) and the orientation about the $x$-axis and the $y$-axis (tilting angles). Computer-aided automated pose estimation methods can in turn also be conditioned to these viewing angles by strictly acquiring training data from the expected working positions. We employ this conditioning in the screw datasets by only using images captured with a rotation in range of $r_{x} \in [-45^{\circ},+45^{\circ}]$, $r_{y} \in  [-45^{\circ},+45^{\circ}]$, $r_{z} \in  [-180^{\circ},+180^{\circ}]$ from the starting position of the X-ray system. Finally, in practice, a projection of the object 3D model will be rendered instead of the bounding-box along with strictly relevant transformation axes to further reduce ambiguities and present clinically-relevant information.
\section{Results}
\label{sec:results}
This section presents the results of the proposed approach on the various datasets used during the development of a method for accurate 6-DoF pose estimation in X-ray.
\begin{figure*}
\centering
\subfloat{{\includegraphics[width=0.24\linewidth]{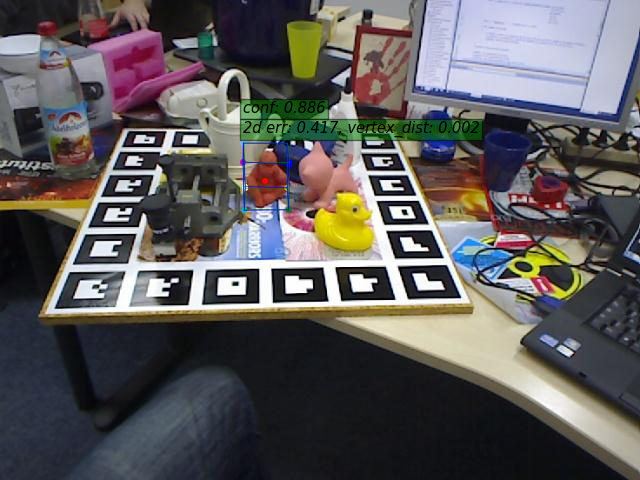}}}\hspace{0.02cm}
\subfloat{{\includegraphics[width=0.24\linewidth]{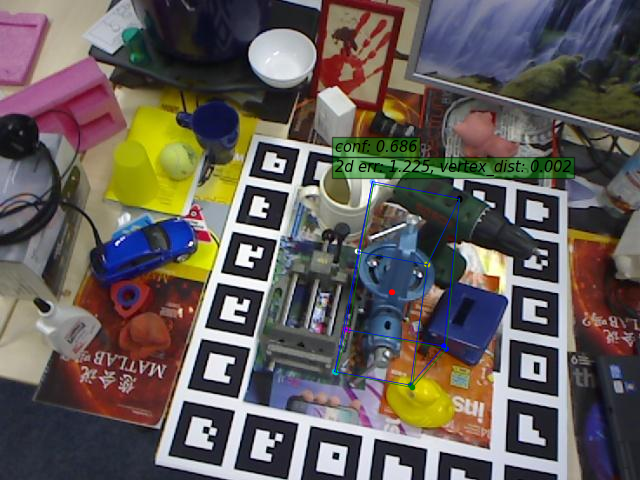}}}\hspace{0.02cm}
\subfloat{{\includegraphics[width=0.24\linewidth]{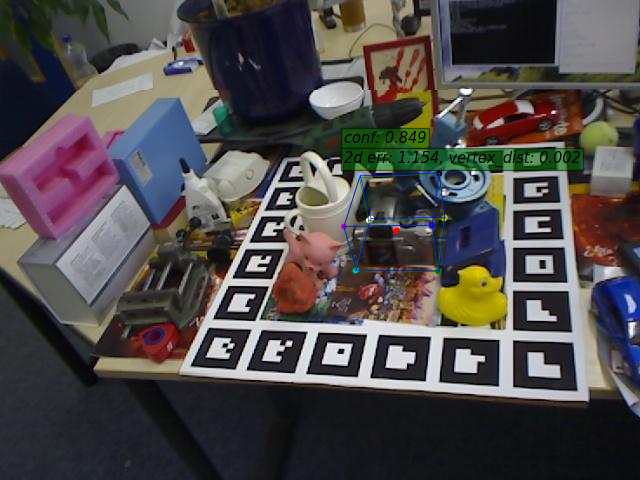}}}\hspace{0.02cm}
\subfloat{{\includegraphics[width=0.24\linewidth]{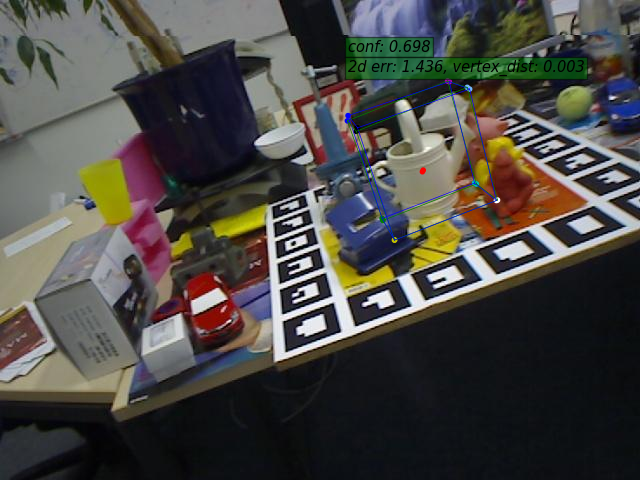}}}
\vspace{-0.25cm}
\subfloat{{\includegraphics[width=0.24\linewidth]{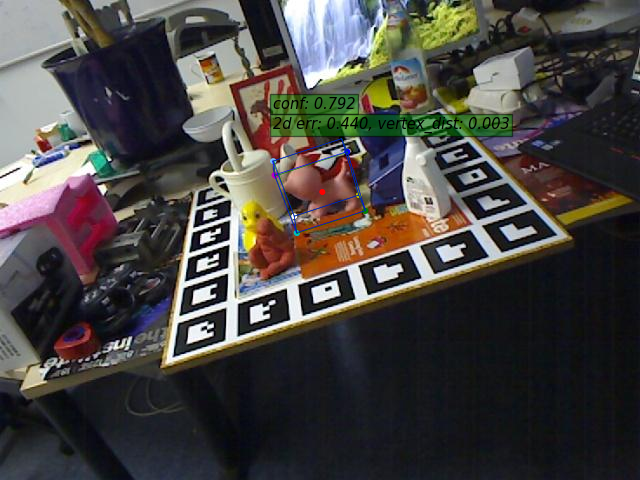}}}\hspace{0.02cm}
\subfloat{{\includegraphics[width=0.24\linewidth]{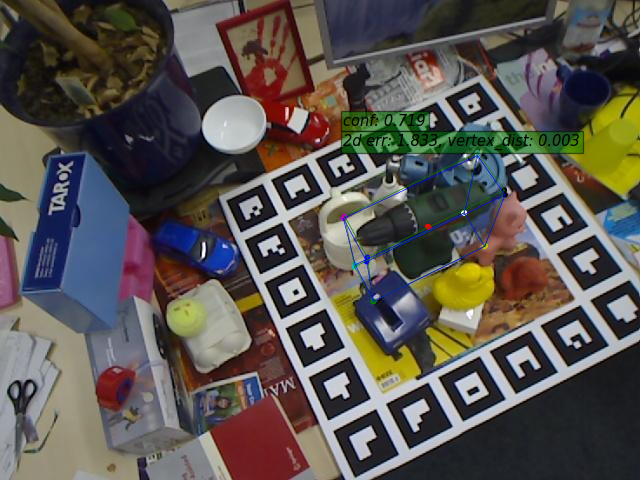}}}\hspace{0.02cm}
\subfloat{{\includegraphics[width=0.24\linewidth]{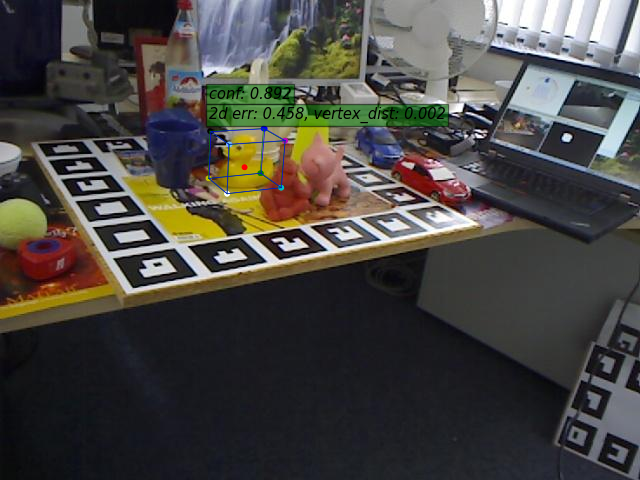}}}\hspace{0.02cm}
\subfloat{{\includegraphics[width=0.24\linewidth]{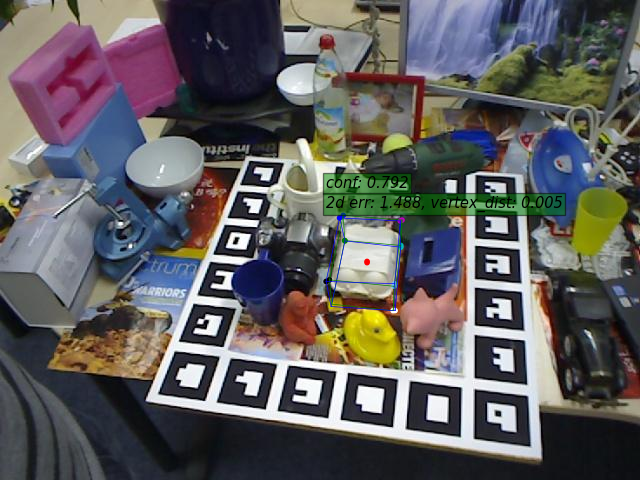}}}
\vspace{-0.25cm}
\subfloat{{\includegraphics[width=0.24\linewidth]{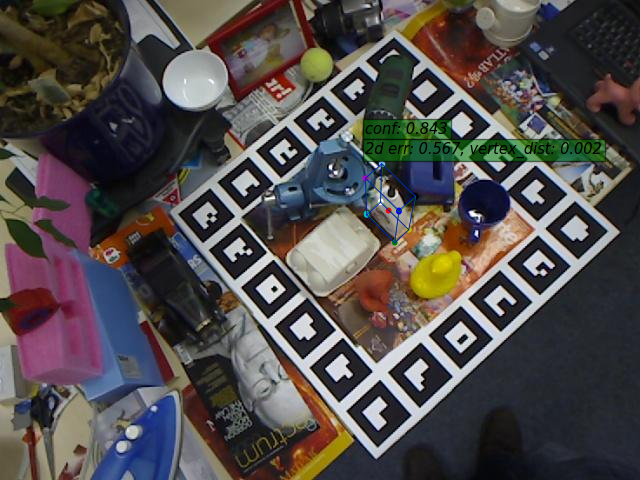}}}\hspace{0.02cm}
\subfloat{{\includegraphics[width=0.24\linewidth]{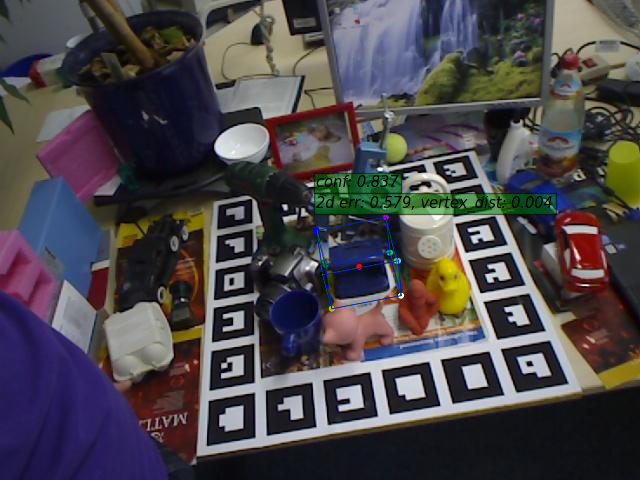}}}\hspace{0.02cm}
\subfloat{{\includegraphics[width=0.24\linewidth]{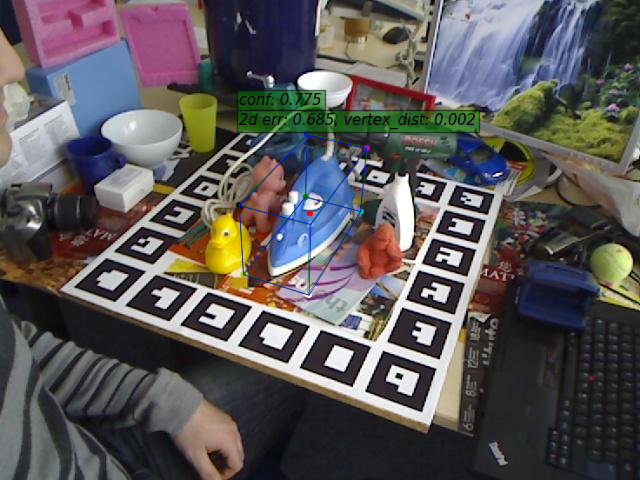}}}\hspace{0.02cm}
\subfloat{{\includegraphics[width=0.24\linewidth]{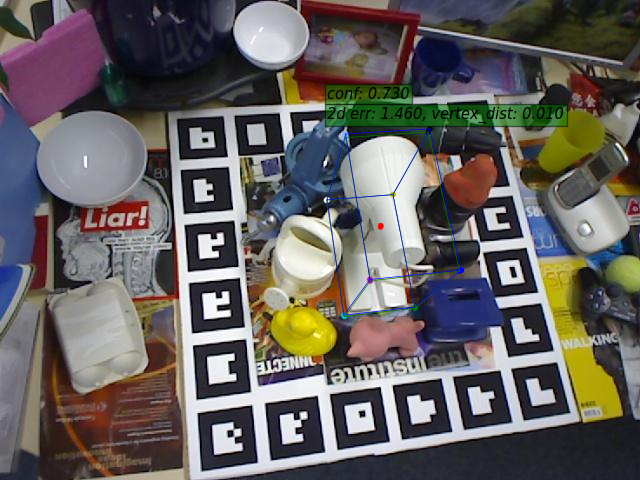}}}
\caption{Example predictions of different objects for qualitative evaluation of the proposed YOLOv5-6D model on the LINEMOD test dataset. Green 3D bounding
boxes visualize ground-truth poses while our estimated poses are represented by blue boxes. The objectiveness, 2D reprojection error and 3D vertices distance are depicted in the floating annotation.}
\label{LineMOD qual}
\end{figure*}
\subsection{Object Pose Estimation in RGB Images}
The quantitative results of the accuracy of our experiments on the LINEMOD dataset are presented in Table~\ref{LineMOD 2D Results} and Table~\ref{LineMOD 3D Results}, while the qualitative results are shown in Figure~\ref{LineMOD qual}. We compare the proposed approach against seven competitive object pose estimation methods on the LINEMOD dataset: YOLO6D~\cite{yolo6D}, PoseCNN~\cite{posecnn}, PVNet~\cite{pvnet}, Gen6D~\cite{gen6d}, EfficientPose~\cite{EfficientPose}, RNNPose~\cite{rnnpose} and EPro-PnPv2~\cite{epro}. The accuracy reported in the respective papers are used for comparison. In addition, Table~\ref{LineMOD 3D Results} and Table~\ref{tab:pose_estimation_properties} categorizes the methods as described in Section~\ref{sec:relatedwork} and based on our findings. The table indicates the type of network employed during the \textbf{1st Stage} of the multi-stage methods and the \textbf{Type} of approach (direct or PnP) utilized for obtaining the object pose.

As can be seen in Table~\ref{LineMOD 2D Results}, YOLOv5-6D realizes an average increase of 9.07\% on the 2D reprojection performance metric over the YOLO6D model. On the ADD(-S) metric (Table~\ref{LineMOD 3D Results}), YOLOv5-6D shows a strong performance increase (40.98\%) over its predecessor and realizes competitive results against SOTA alternative methods, while being much faster (see Table~\ref{fig:speed_tradeoff}). Comparisons to the seven SOTA alternative methods are added to set strong baselines and for completeness, as we evaluate YOLOv5-6D as a new architecture in general.
\begin{table}[hb]
\centering
\caption{Comparison of on LINEMOD in terms of the 2D reprojection metric.}
\setlength{\tabcolsep}{3pt}
\begin{tabular}{c|cc}
\toprule
Object & YOLO6D  & YOLOv5-6D\\
\midrule
Ape & 92.10 & 99.24\\
Benchvise & 95.06 &  99.61\\
Cam & 93.14 &  99.71\\
Can & 97.44 & 99.80\\
Cat & 97.41 &  99.80\\
Driller & 79.41  & 98.61\\
Duck & 94.65 & 99.16\\
Eggbox & 90.33 & 99.34\\
Glue & 96.53 & 99.61\\
Holepuncher & 92.86 &  99.91\\
Iron & 82.94 & 99.59\\
Lamp & 76.87 &  98.85\\
Phone & 86.07 &  99.52\\
\midrule
Average & 90.37 & 99.44\\
\bottomrule
\end{tabular}
\label{LineMOD 2D Results}
\end{table}

\begin{table*}[ht]
\centering
\caption{ Comparison of the proposed approach with alternative methods on LINEMOD using the 10\% ADD and ADD-S(*) metric.}
\setlength{\tabcolsep}{3pt}
\begin{tabular}{c | c|c|c|c|c|cc|c|c}
\toprule

\bf{Method} & YOLO6D & PoseCNN & PVNet &  Gen6D (Model Free) & EfficientPose & \multicolumn{2}{c|}{RNNPose} & EPro-PnPv2 & YOLOv5-6D (Ours)\\
\midrule
\bf{Type} & PnP & Direct & PnP  &  Direct Volume Refine & Direct & \multicolumn{2}{c|}{PnP Refinement} & PnP  & PnP\\
\bf{1st. Stage}       &        &    &     &  Det. \& View point select   &               & \multicolumn{1}{c|}{PoseCNN} &   PVNet &    Faster-RCNN Det.  & \\
\midrule
\multicolumn{1}{c}{\bf{Object}}\\
\midrule
Ape & 21.62 & 25.62 & 43.62 & - & 87.71  & \bf{88.19} & 85.62 & - &  87.81\\
Benchvise & 81.80 & 77.11 & 99.90 & 77.03 & 99.71  & \bf{100.0} & \bf{100.0}  & - &  \bf{100.0}\\
Camera & 36.57 & 47.25 & 86.86 & 66.67 & 97.94  & 98.04 & \bf{98.43} & -  & 97.45\\
Can & 68.80 & 69.98 & 95.47  & - & 98.52 &  99.31 & \bf{99.51} & - & 99.31\\
Cat & 41.82 & 56.09 & 79.34 & 60.68 &\bf{98.00} & 96.41 & 96.41 & -  & 96.21\\
Driller & 63.51 & 64.92 & 96.43 & 67.39 &\bf{99.90} & 99.70 & 99.50 & - & 99.11\\
Duck & 27.23 & 41.74 & 52.58 & 40.47 &\bf{90.00} & 89.30 & 89.67 & -  & 86.57\\
Eggbox* & 69.58 & 98.50 & 99.15 & - &\bf{100.0} &  99.53 & \bf{100.0}& -  & \bf{100.0}\\
Glue* & 80.02 & 94.98 & 95.66 & - & \bf{100.0} & 99.71 & 97.30 & -  & \bf{100.0}\\
Holepuncher & 52.24 & 42.63 & 81.92 & - & 95.15 & \bf{97.43} & 97.15 & -  & 95.34\\
Iron & 74.97 & 70.17 & 98.88 & - & 99.69 &  \bf{100.0} & \bf{100.0} &-  & 99.19\\
Lamp & 71.11 & 70.73 & 99.33 & 89.83 &\bf{100.0} & 99.81 &\bf{100.0} & -  & \bf{100.0}\\
Phone & 47.74 & 53.07 & 92.41 & - & 97.98 & 98.39 & \bf{98.68} & -  & 97.89\\
\midrule
Average & 55.95 & 63.26 & 86.27 & 67.01 & 97.35 & \bf{97.37} & 97.10 & 96.36  & 96.84\\
\bottomrule
\end{tabular}
\label{LineMOD 3D Results}
\end{table*}

\subsection{Inference Time}\label{sec:inference_time}
\begin{figure}[b]
\centering
\includegraphics[width=\linewidth]{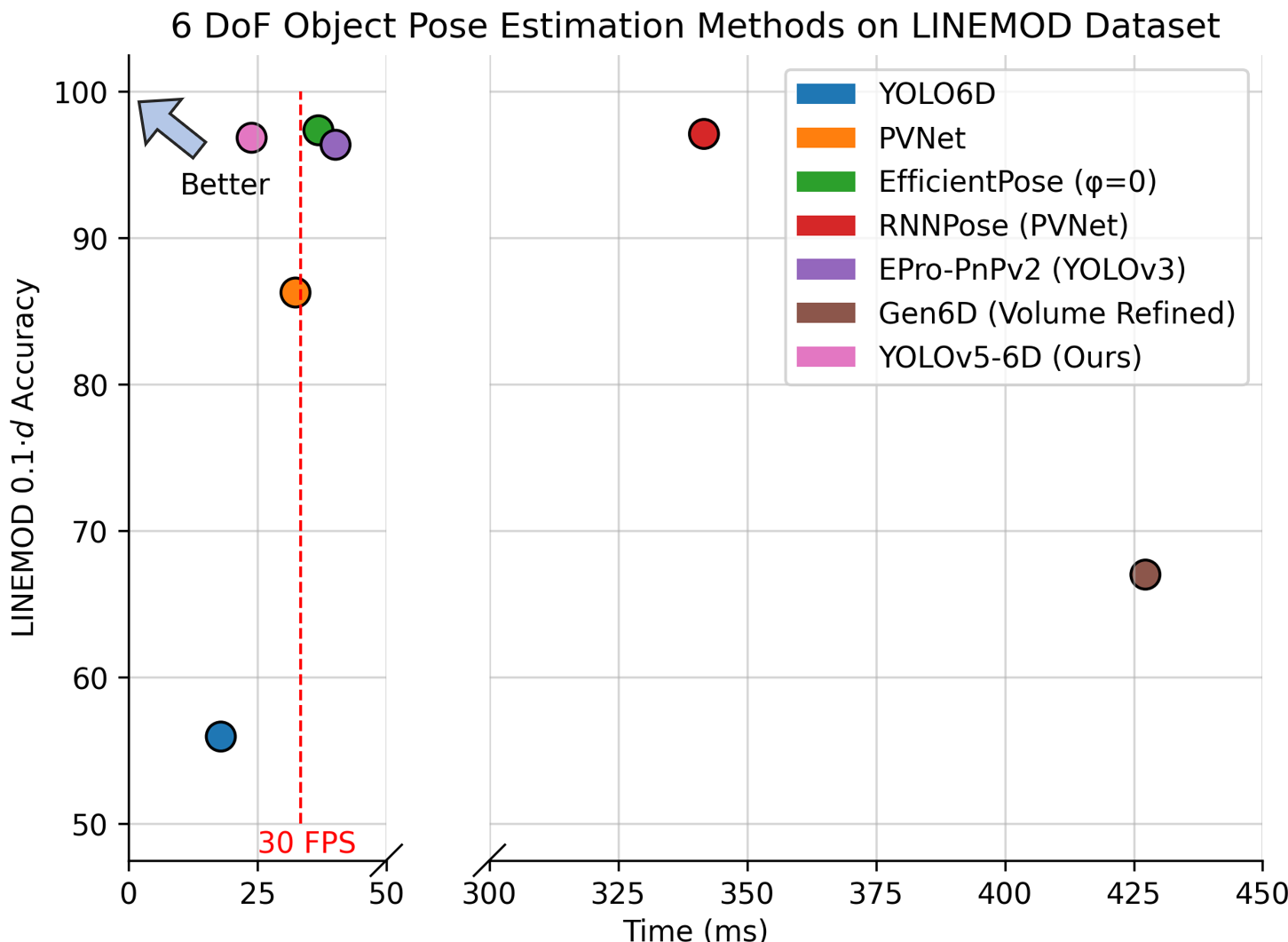}
\caption{Accuracy and inference-time comparison of YOLOv5-6D and competitive alternative methods. Measurements are obtained with a unity batch size.}
\label{fig:speed_tradeoff}
\end{figure}

To assess the real-time performance of the proposed YOLOv5-6D model, aimed at achieving 30~FPS, we have conducted a comparative analysis of its inference time against other leading 6-DoF pose estimation methods. This comparison is carried out under uniform hardware conditions (see Section~\ref{sec:training_details}) to ensure fairness, unless otherwise stated. We have employed each method as described in the respective research papers and as made publicly available. In all cases, we use the LINEMOD cat dataset (640$\times$480$\times$3~images) with corresponding pre-trained weights for the cat object, with the exception of GEN6D (no object specific model is required). The text below summarizes our findings.
\begin{itemize}
    \item \textbf{YOLO6D}: Achieves a total inference time of 17.9~ms per frame (55.8~FPS), which is 5~FPS faster than originally reported. This includes image loading to GPU (0.6~ms), model forward pass (4.5~ms), and filtering the predictions (12.8~ms).
    \item \textbf{PVNet}: Yields an inference time of 32.4~ms (30.9~FPS), encompassing data loading time (3.5~ms), PVNet model forward pass (17.6~ms), and the RANSAC-based voting scheme (11.3~ms) used to obtain the reported accuracy.
    \item \textbf{EfficientPose($\phi=0$)}: Demonstrates a pose prediction time of 36.8~ms (27.15~FPS), comparable to the reported 27.45~FPS. This includes data preprocessing (14.1~ms) and model inference (22.7~ms).
    \item \textbf{RNNPose}: Utilizes initial poses from PVNet (no end-to-end solution is developed), with refinement inference time depending on the number of recurrent and rendering cycles employed. Each recurrent iteration involves correspondence field (CF) Estimation (11.0~ms), pose optimization (0.4~ms), CF rectification (4.4~ms) for a total execution time of 15.8~ms. The rendering cycle includes reference image rendering (9.4~ms), 3D feature rendering (3.6~ms), image feature encoding (2.6~ms), followed by the earlier mentioned recurrent iterations. As depicted in the RNNPose paper (Figure~5 \& Table~2), approximately four rendering cycles (4~total) with each running four recurrent iterations (16~total) are required to obtain the SOTA LINEMOD performance reported in the paper. In addition to the rendering cycles modules (15.6$\times$4=62.5~ms) and recurrent iterations (15.8$\times$16=253.0~ms), a once-off data loading time (2.9~ms) and running the 2D-3D Hybrid Net (2.1~ms) bring the refinement execution time to 320.5~ms and the total execution time (with the addition of PVNet initial poses without RANSAC voting) to 341.6~ms (2.93~FPS). 
    \item \textbf{EPro-PnP}: Exhibits a rapid inference time of 10.1~ms (98.5~FPS), requiring 0.3~ms for image-crop data loading, 4.8~ms for model forward pass, 1.1~ms for postprocessing, and 3.9~ms for the PnP calculation. However, the 2-staged approach requires an earlier model to detect the objects of interest and provide exact crops to the EPro-PnP part. The research largely improves on the earlier work of Li~\textit{et al.} called CDPN~\cite{li2019cdpn}, which utilizes the same 2-stage approach. While EPro-PnPv2 employs Faster-RCNN~\cite{ren2015fasterrcnn}, a relatively old and slower object detector, no investigation has been conducted into how the model performs based on the provided input crop. Alternatively, in the CDPN approach (Table~3 \& 4) the authors show that by using YOLOv3 they get slightly lower performance (ADD(-S) 89.80 with YOLOv3 vs. ADD(-S) 89.86 with Faster-RCNN), but with a significant speed improvement (30~ms vs 76~ms). Since neither a detection implementation is discussed nor provided along with EPro-PnPv2, we have employed the YOLOv3-based performance reported in CDPN~\cite{li2019cdpn} in our comparison. This enables a total EPro-PnPv2-based pose prediction in 40.2~ms (24.9~FPS).
    \item \textbf{Gen6D}: A 3D object model-free and generic estimator, achieves a novel object pose prediction at 427.26~ms per frame (2.34~FPS), including initial object detection (125.7~ms), viewpoint selection (37.1~ms), and pose refinement (3$\times$88.0~ms=256.1~ms).
    \item \textbf{YOLOv5-6D}: Realizes single-shot object 6-DoF pose estimation at 41.88~FPS (inference time of 23.88~ms per frame). Table~\ref{tab:inference time} depicts the exact execution time per module of the YOLOv5-6D model for both the LINEMOD and our X-ray datasets.
\end{itemize}
For a comprehensive comparison we include Figure~\ref{fig:speed_tradeoff}, illustrating the speed versus average accuracy trade-off on the LINEMOD dataset. The proposed YOLOv5-6D enables single-shot, real-time object 6-DoF pose estimation, demonstrates its efficacy on both the LINEMOD and our larger X-ray datasets. This comparison highlights the balance between speed and accuracy of 6-DoF pose estimation methods and underscores the efficiency of our proposed model. Finally, the results of all findings are summarized in Table~\ref{tab:pose_estimation_properties}.

\begin{table}
\centering
\caption{YOLOv5-6D inference time on the LINEMOD and X-ray datasets. Measurements are obtained with a unity batch size.}
\setlength{\tabcolsep}{5pt}
\begin{tabular}{l | cc}
\toprule
Operation & LINEMOD & X-ray \\
Image size &  640$\times$480$\times$3 & 960$\times$742$\times$1 \\
\midrule
 Tensor to cuda &  0.22 ms & 0.20 ms \\
 Predict & 23.03 ms & 29.82 ms\\
 Filter predictions & 0.52 ms & 0.42 ms \\
 ePnP & 0.11 ms & 0.07 ms \\
 \midrule
 Total time& 23.88 ms  & 30.51 ms\\
   Frame rate (FPS) & 41.88  & 32.78 \\
\bottomrule
\end{tabular}
\label{tab:inference time}
\end{table}
\subsection{X-Ray Pose Estimation}
\begin{figure}[t]
\includegraphics[width=0.98\linewidth, trim={7.20cm 0cm 14.0cm 0.0cm},clip]{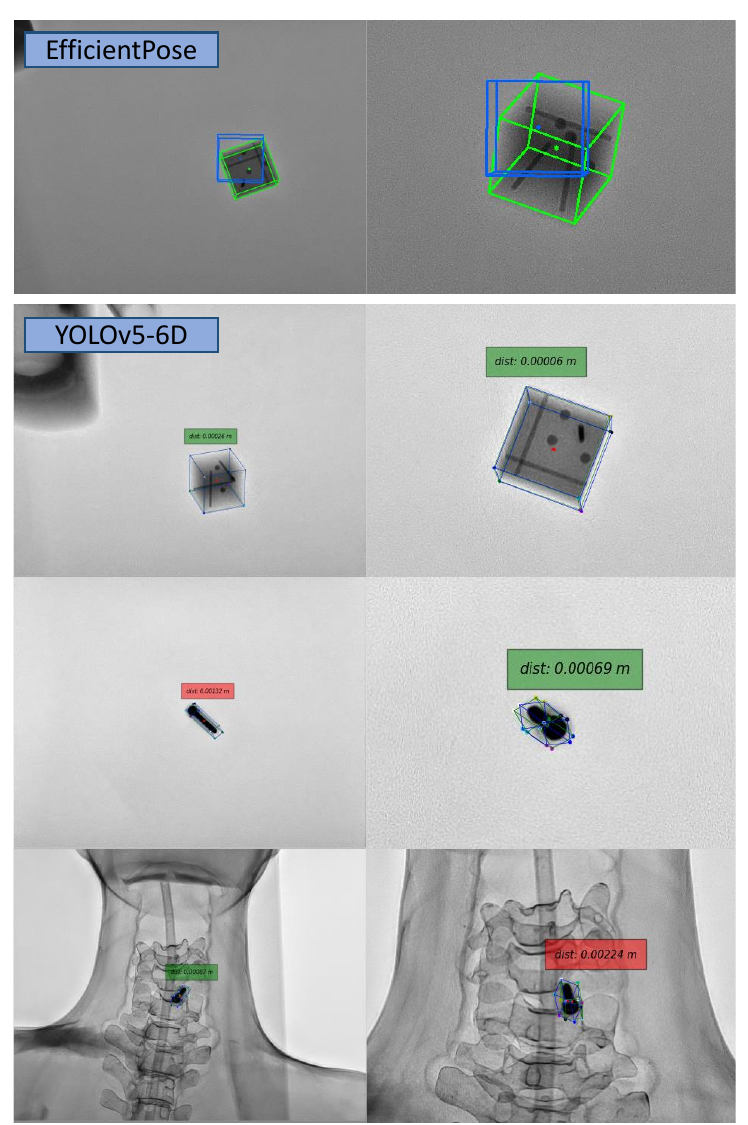}
\caption[x-ray Predictions]{Example predictions for qualitative evaluation on the Cube Validation, Screw Validation and Screw Phantom Test datasets. The first column presents images in the original resolution and the second highlight a zoomed area. Green 3D bounding boxes visualize ground-truth poses, while our estimated poses are represented by blue boxes. The average 3D vertices distance for YOLOv5-6d is shown in the floating text box with a color representing pass (green) or fail (red) by the 1-mm ADD(-S) metric. Images are rendered with the respective code bases.}
\label{fig:X-ray results}
\end{figure}

In Section~\ref{sec:introduction} and~\ref{sec:X-ray Pose Estimation} we present the hard requirements for an object 6-DoF pose estimation method to be successful in the medical X-ray domain. In summary, the method needs to be (1)~very accurate, (2)~incorporate image acquisition geometry and (3)~be fast to enable real-time analysis. Given these strict requirements and the analysis of the results of the various methods on the LINEMOD dataset, YOLOv5-6D presents itself as the only viable candidate in our X-ray domain. To test this assertion, we conduct experimental analysis of YOLOv5-6D and EfficientPose($\phi=0$) in the X-ray domain. The quantitative results of the experiments on the X-ray datasets can be observed in Table~\ref{X-ray Results} and corresponding qualitative results are shown in Figure~\ref{fig:X-ray results}.
\begin{table}[b]
\setlength{\tabcolsep}{4pt}
\centering
\caption{Performance of YOLOv5-6D and EfficientPose on the X-ray Cube and X-ray Screws datasets at various distance thresholds. Note the use of distance $d$ as a parameter which is 30~\MakeLowercase{mm} for the Cube and 34.3\MakeLowercase{mm} for the Screw.}
\begin{tabular}{l|c|ccc}

\toprule
Model & Eff.Pose &\multicolumn{3}{c}{YOLOv5-6D} \\
Metric [mm] & Cube Val& Cube Val& Screw Val& Screw Test\\
\midrule
 ADD(-S) 0.1$\cdot d$ & 0.0 &  99.27 & 96.87 & 92.41\\
 ADD(-S) 0.05$\cdot d$ & 0.0 &  97.08 & 87.50 & 81.01\\
 ADD(-S) 1.0 mm  & 0.0 & 93.43 & 75.0 & 55.70\\
 ADD(-S) 0.02$\cdot d$ & 0.0 & 82.48 & 65.62 & 43.04\\
 \midrule
 3D Transl.err. [mm] &  13.8 $\pm$ 4.5 & 0.35$\pm$0.21& 0.82$\pm$0.43 & 1.27$\pm$0.47\\
 3D Ang.err. [deg.] & 33.7 $\pm$ 8.3 & 1.45$\pm$1.29 & 3.18$\pm$1.72 & 3.79$\pm$2.72\\
\bottomrule
\end{tabular}
\label{X-ray Results}
\end{table}

The conducted experiments show that the YOLOv5-6D model can predict relevant 2D key points for accurate 6-DoF pose estimation, notably also in challenging scenarios like our X-ray datasets where the focal length varies by up to 28 cm and the object undergoes translation and rotation. In contrast, EfficientPose, tends to converge to a mean pose present in our dataset, reflecting its low performance in such settings. This is expected due to the ambiguity present in the pose if the method does not have access to camera intrinsic parameters during training.

Specifically, for our X-ray datasets featuring two small instruments, the YOLOv5-6D model achieves a high accuracy of 99.27\% for the asymmetrical cube and 96.41\% for the symmetrical bone screw. At a 1-mm distance threshold, the pose of the asymmetrical cube is estimated with a 93.43\% accuracy. Similarly, at a 1-mm threshold, the pose of the symmetrical bone screw is accurately acquired in 75\% of the validation cases. The same model trained on images only containing the screw (and heavy augmentation) is then applied to the screw and the spine phantom dataset. The model shows comparable and high accuracy at the 0.1$\cdot d$ (3.43~mm) and 0.05$\cdot d$ (1.72~mm) threshold, but experiences a large drop in performance at the smaller distances. We do not report the 2D reprojection error in the X-ray datasets, because the symmetry around the $z$-axis of the screw allows for multiple plausible 2D key point predictions that will resolve the correct object pose. This is visually proven and illustrated in Figure~\ref{fig:X-ray results}. Lastly, Table~\ref{tab:inference time} depicts the inference time of the YOLOv5-6D model for the two domains. 
\section{Discussion}
\label{sec:discussion}

A novel YOLOv5-6D method for accurate 2D key point prediction and associated pose estimation is developed, while considering the image acquisition geometry. Prior key point-based methods are not fast or accurate enough for real-world application, especially in the medical domain where accuracy is essential. The YOLOv5-6D model builds on advancements in the YOLO object detection series to improve prediction accuracy of 2D key points and enable correct pose estimation through solving the 2D/3D object bounding-box correspondences with PnP. In addition, we have presented a new data capturing method for 6-DoF tasks that utilizes an optical camera attached to the X-ray detector. The approach allows for data acquisition across all X-ray geometries and objects without adding image artifacts (such as AruCo markers or calibration domes) to the final X-ray image, or relying on accurate X-ray system sensors to acquire object-pose labels. 
\begin{table}
\centering
\caption{Pose estimation method properties based on LINEMOD}
\setlength{\tabcolsep}{4pt}
\begin{tabular}{r| cccc}
\toprule
Method & Task-agnostic & Cam. Intrinsic & Real-Time & $\geq 90$ Acc.\\
\midrule
YOLO6D & - & \checkmark & \checkmark & - \\ 
PoseCNN & - & - & - &-\\
PVNet & - & \checkmark & \checkmark & - \\ 
Gen6D  & \checkmark & - & - & -\\
EfficientPose & - & - & - & \checkmark\\
RNNPose & - & \checkmark & - & \checkmark\\
EPro-PnPv2 & - & \checkmark & - & \checkmark\\
YOLOv5-6D & - & \checkmark & \checkmark & \checkmark\\

\bottomrule
\end{tabular}
\label{tab:pose_estimation_properties}
\end{table}
The YOLOv5-6D method generalizes across domains and imaging systems. This generalization is evident from (1)~its application to the RGB images, (2)~X-ray images obtained with different acquisition geometries and (3)~different levels of semantic complexity in the X-ray image contents.

With respect to the first~(1) aspect of generalization, the YOLOv5-6D model shows competitive results on the public LINEMOD RGB dataset with an average ADD(-S) score of 96.84\% compared to the current SOTA (RNNPose~\cite{rnnpose}) with 97.37\%, as demonstrated in Table~\ref{LineMOD 3D Results}. However, the proposed method is considerably faster (41.88~FPS vs 27.15~FPS of EfficientPose) in execution at this level of accuracy and leverages the imaging geometry as summarized in Figure~\ref{fig:speed_tradeoff} and Table~\ref{tab:pose_estimation_properties}. These attributes make the method appealing for real-time instrument pose estimation in the X-ray domain. 

The second~(2) aspect is addressing generalization towards different X-ray geometry. Here, images are obtained with various hardware configurations using higher input image resolution. The images are without depth information and typically contain low contrast of the objects of interest. The proposed YOLOv5-6D successfully predicts relevant 2D bounding-box key points for both X-ray objects included in this research, enabling highly accurate pose estimation (Table~\ref{X-ray Results}) with a translation error of only 0.35~mm$\pm$0.21.

The third aspect~(3) is about generalizing across different levels of semantic complexity. The model trained for screw pose estimation in a simple training environment generalizes to the new and more clinically-relevant domain containing the human phantom. This generalization is evident from Table~\ref{X-ray Results}. Most notably, the proposed approach is able to accurately estimate the pose of a small cannulated cancellous bone screw up to an impressive 75.0\% at 1~mm by the ADD(-S) metric on the validation set. In addition, the same model generalizes well to the new and more complex test set containing a spine phantom. Here, we observe a similarly high 92.41\% ADD(-S) score at 0.1$\cdot d$. As a very hard final test, we evaluate the pose accuracy at 1 mm ADD(-S) which shows a drop to 55.70\% in comparison to the validation set. This performance drop is expected due to the stringency of the test, but we consider that it can rather be traced to inaccuracy in the labels. For example, prior to the labeling process, the offset from the instrument to the ChArUco frame is manually and precisely measured. However, with the screw being placed in a spine phantom, this measurement becomes considerably more difficult and error prone. Our measurements for acquiring the ground-truth labels of the test set are likely to be off by $\pm$1~mm, which results in a lower performance of the proposed model at these distances. The performance of the proposed method on these 4 datasets expresses the generalization ability of the method and future research can further elaborate by testing on other instruments.

This research is one of the first to propose tracking the actual screw instead of the surgical path or screw placement tools for assisted clinical guidance. The YOLOv5-6D method enables accurate and fast 6-DoF pose estimation of the screw with respect to the X-ray detector or source. By combining this method with a spine tracking system, such as the one proposed by Manni~\textit{et al.}~\cite{Manni}, the screw pose can be determined with respect to the target location on the spine, enabling precise screw placement and its validation without the need for postoperative CT. 
\section{Future Work \& Limitations}
\label{sec:limitations}
In our study, we have focused on single-object and single-class pose estimation and have not collected data to investigate multi-object and multi-class pose estimation. However, simultaneous multi-object pose estimation is an intriguing area of future research in the context of spinal screw placement, as multiple screws are typically used in this procedure. We conjecture that the proposed YOLOv5-6D model can be leveraged for estimating the pose of multiple similar-sized screws without the need to retrain the model. In scenarios where multiple identical screws of different sizes (multi-class) are used, the ill-posed nature of X-ray imaging may hinder the ability to distinguish between these objects. Nonetheless, it is worth noting that the screws used during the clinical procedure are known beforehand, and the corresponding 3D screw model can be manually linked to accurately determine the pose of each screw, regardless of its size. This presents an exciting opportunity for future research to explore the feasibility and effectiveness of this approach in the context of spinal screw placement.

Furthermore, we show that the model successfully estimates the screw pose outside of its training distribution. However, it is still evaluated in a rather simple context and future work needs to fully explore the limitations of the approach under more clinically relevant conditions.

While we investigate object pose estimation under variable X-ray imaging geometry, similar challenges arise outside of the medical domain, such as optical cameras with adjustable focal lengths used for zooming. In these cases, failure to account for the change in imaging geometry can lead to incorrect object pose estimations from the captured image. Moreover, satellite pose estimation~\cite{kisantal2020satellitechallange, satalitelightweight} is a domain that already faces this challenge and can straightforwardly benefit from the proposed approach.
\section{Conclusion}
\label{sec:conclusion}
This research presents a novel YOLOv5-6D model for accurate 6-DoF instrument pose estimation to provide clinical guidance under varying image acquisition geometries of X-ray systems. The YOLOv5-6D model utilizes recent advancements made in object detection to improve the accuracy of pose estimation. The model performance is first established on public benchmarks and achieves competitive results on the LINEMOD dataset with an average 0.1$\cdot d$ ADD(-S) score of 96.84\%, while being considerably faster up to 42~FPS, than existing approaches with comparable levels of accuracy. Using the newly proposed data capturing method, two new X-ray datasets are constructed, consisting of a calibration cube and a clinically relevant cancellous bone screw. The model shows strong results in estimating the pose of these instruments in the ambiguous X-ray domain as indicated by the high 0.1$\cdot d$ ADD(-S) 92.41\% performance. Generalization of the model trained for screw pose estimation is evaluated on a spine-phantom test set and achieves compelling results. These results demonstrate that the proposed approach has a strong potential to effectively assist clinicians in instrument maneuvering and placement during minimally invasive surgeries. It has also been substantiated that the model can be generalized in terms of the optical sensing, X-ray image pose estimation with different acquisition geometries whilst addressing higher levels of semantic complexity in the image contents.

\section*{Acknowledgment}
The authors would like to thank Philips Image Guided Therapy Systems R\&D for the support and funding of this research. The access and assistance with using the IGT X-ray system is greatly appreciated.  

\bibliography{main} 
\bibliographystyle{IEEEtran}

\end{document}